\documentclass[sn-basic,iicol]{sn-jnl}

\usepackage{graphicx}
\usepackage{multirow}
\usepackage{amsmath,amssymb,amsfonts}
\usepackage{amsthm}
\usepackage{mathrsfs}
\usepackage[title]{appendix}
\usepackage{xcolor}
\usepackage{textcomp}
\usepackage{manyfoot}
\usepackage{booktabs}
\usepackage{algorithm}
\usepackage{algorithmicx}
\usepackage{algpseudocode}
\usepackage{listings}
\usepackage{epsfig}
\usepackage{makecell}
\usepackage{float}
\usepackage{subcaption}
\usepackage{pifont}
\usepackage{enumitem}
\usepackage{wrapfig}
\usepackage{ragged2e}
\usepackage{blindtext}
\usepackage{multicol}
\usepackage{cellspace, tabularx}

\newcommand{\xmark}{\ding{55}}

\def\eg{\emph{e.g}.} 
\def\ie{\emph{i.e}.}

\begin{document}

\title{Swap Attention in Spatiotemporal Diffusions for Text-to-Video Generation}

\author[1]{Wenjing Wang}\email{daooshee@pku.edu.cn}
\author*[2]{Huan Yang}\email{huayan@microsoft.com}
\author[2]{Zixi Tuo}\email{v-zixituo@microsoft.com}
\author[2]{Huiguo He}\email{v-huiguohe@microsoft.com}
\author[2]{Junchen Zhu}\email{v-junchenzhu@microsoft.com}
\author[2]{Jianlong Fu}\email{jianf@microsoft.com}
\author[1]{Jiaying Liu}\email{liujiaying@pku.edu.cn}

\affil[1]{\orgdiv{Wangxuan Institute of Computer Technology}, \orgname{Peking University}}
\affil[2]{\orgname{Microsoft Research Asia}}

\abstract{With the explosive popularity of AI-generated content (AIGC), video generation has recently received a lot of attention. Generating videos guided by text instructions poses significant challenges, such as modeling the complex relationship between space and time, and the lack of large-scale text-video paired data. Existing text-video datasets suffer from limitations in both content quality and scale, or they are not open-source, rendering them inaccessible for study and use. For model design, previous approaches extend pretrained text-to-image generation models by adding temporal 1D convolution/attention modules for video generation. However, these approaches overlook the importance of jointly modeling space and time, inevitably leading to temporal distortions and misalignment between texts and videos. In this paper, we propose a novel approach that strengthens the interaction between spatial and temporal perceptions. In particular, we utilize a swapped cross-attention mechanism in 3D windows that alternates the ``query'' role between spatial and temporal blocks, enabling mutual reinforcement for each other. Moreover, to fully unlock model capabilities for high-quality video generation and promote the development of the field, we curate a large-scale and open-source video dataset called HD-VG-130M. This dataset comprises 130 million text-video pairs from the open-domain, ensuring high-definition, widescreen and watermark-free characters. A smaller-scale yet more meticulously cleaned subset further enhances the data quality, aiding models in achieving superior performance. Experimental quantitative and qualitative results demonstrate the superiority of our approach in terms of per-frame quality, temporal correlation, and text-video alignment, with clear margins.}


\maketitle

\section{Introduction}
\label{sec:introduction}

Automated video production is experiencing a surge in demand across various industries, including media, gaming, film, and television~\citep{2017_DigiPro,2021_Playable}.
This increased demand has propelled video generation research to the forefront of deep generative modeling, leading to rapid advancements in the field~\citep{2022_VDM, 2016_VG_ICLR, 2017TGAN,2018MoCoGAN, 2016_VG_NIPS}.
In recent years, diffusion models~\citep{2020DDPM} have demonstrated remarkable success in generating visually appealing images in open-domains~\citep{2022LDM,DALLEE2,SDXL,SD3}.
Building upon such success, in this paper, we take one step further and aim to extend their capabilities to high-quality text-to-video generation.

As is widely known, the development of open-domain text-to-video models poses grand challenges, due to the limited availability of large-scale text-video paired data and the complexity of constructing space-time models from scratch.
To solve the challenges, current approaches are primarily built on pretrained image generation models.
These approaches typically adopt space-time separable architectures, where spatial operations are inherited from the image generation model~\citep{2022_VDM,2023CogVideo}.
To further incorporate temporal modeling, various strategies have been employed, including pseudo-3D modules~\citep{MakeAVideo,zhou2022magicvideo}, serial 2D and 1D blocks~\citep{VideoLDM,2022ImagenVideo}, and parameter-free techniques like temporal shift~\citep{2023LatentShift} or tailored spatiotemporal attention~\citep{wu2022tuneavideo,Text2VideoZero}. 
However, these approaches overlook the crucial interplay between time and space for visually engaging text-to-video generation.
On one hand, parameter-free approaches rely on manually designed rules that fail to capture the intrinsic nature of videos and often lead to the generation of unnatural motions.
On the other hand, learnable 2D+1D modules and blocks primarily focus on temporal modeling, either directly feeding temporal features to spatial features, or combining them through simplistic element-wise additions.
This limited interactivity usually results in temporal distortions and discrepancies between the input texts and the generated videos, thereby hindering the overall quality and coherence of the generated content.

To address the above issues, we take one step further in this paper which highlights the complementary nature of both spatial and temporal features in videos.
Specifically, we propose a novel Swapped spatiotemporal Cross-Attention (Swap-CA) for text-to-video generation. 
Instead of solely relying on separable 2D+1D self-attention~\citep{2021_space_time_attn} or 3D window self-attention~\citep{2022VideoSwin} that replace computationally expensive 3D self-attention, we aim to further enhance the interaction between spatial and temporal features.
Our swap attention mechanism facilitates bidirectional guidance between spatial and temporal features by considering one feature as the query and the other as the key/value.
To ensure the reciprocity of information flow, we also swap the role of the ``query" in adjacent layers.

By deeply interplaying spatial and temporal features through the proposed swap attention, we present a holistic VideoFactory framework for text-to-video generation. In particular, we adopt the latent diffusion framework and design a spatiotemporal U-Net for 3D noise prediction. To unlock the full potential of the proposed model and fulfill high-quality video generation, we construct {a large} video generation dataset, named HD-VG-130M. This dataset consists of 130 million text-video pairs from open-domains, encompassing high-definition, widescreen, and watermark-free characters. 
{We conduct additional data processing, taking into account text, motion, and aesthetics, to create a higher-quality subset. This subset has been shown to effectively enhance video generation performance further.}
Additionally, our spatial super-resolution model can effectively upsample videos to a resolution of $1376\times768$, thus ensuring engaging visual experience. We conduct comprehensive experiments and show that our approach outperforms existing methods in terms of both quantitative and qualitative comparisons. 
In summary, our paper makes the following significant contributions: 

\vspace{2mm}
\begin{itemize}
    \item[-] We reveal the significance of learning joint spatial and temporal features for video generation, and introduce a novel Swapped spatiotemporal Cross-Attention (Swap-CA) mechanism to reinforce both space and time interactions.
    It significantly improves the generation quality, while ensuring precisely semantic alignment between the input text and the generated videos.
    \vspace{1mm}
    \item[-] We curate {\textbf{the first open-source}}\footnote{Project: \url{https://github.com/daooshee/HD-VG-130M}. Our dataset was released in Jan. 2024.} dataset comprising 130 million text-video pairs to-date, supporting high-quality video generation with high-definition, widescreen, and watermark-free characters.
    We proceed with additional processing to extract a higher quality subset and delve into the impact of data processing on video generation.
    We believe this dataset and corresponding analysis will greatly benefit fellow researchers and advance the field of video generation.
\end{itemize}

\vspace{2mm}
The remainder of this paper is organized as follows.
Section \ref{sec:related_works} provides a brief overview of related works.
Section \ref{sec:hd_vg} introduces our proposed HD-VG-130M dataset, analyzes its properties, and introduces the process of constructing a higher-quality subset.
Section \ref{sec:method} presents the proposed text-to-video generation model Video Factory and the Swap-CA design.
Experimental results and concluding remarks are provided in Sections \ref{sec:experiments} and \ref{sec:conclusion}, respectively.

\section{Related Works}
\label{sec:related_works}

\subsection{Text-to-Image Generation}
Generating realistic images from corresponding descriptions combines the challenging components of language modeling and image generation.
Traditional text-to-image generation methods~\citep{2016T2I,ICML_16,AttnGAN,StackGAN} are mainly based on Generative Adversarial Networks (GANs)~\citep{GAN} and are only able to model simple scenes such as birds~\citep{WahCUB_200_2011}.
Later work~\citep{dalle,CogView} extends the scope of text-to-image generation to open domains with better modeling techniques and training data on much larger scales.
In recent years, diffusion models have shown great ability in visual generation~\citep{2021ADM}.
For text-to-image multi-modality generation, GLIDE~\citep{2021GLIDE}, Imagen~\citep{imagen}, DALL·E series~\citep{DALLEE2,DALLEE3}, and Stable Diffusion series~\citep{2022LDM,SDXL,SD3} leverage diffusion models to achieve impressive results. 
Based on these successes, some work extends customization~\citep{2022DreamBooth}, image guidance~\citep{2023PaintbyExample,2023ControlNet}, and precise control~\citep{2022eDiff-I}.
This paper further extends diffusion models for video generation.

\subsection{Text-to-Video Generation.}
Additional controls are often added to make the generated videos more responsive to demand~\citep{2016_VG_ICLR,2017Create_What_You_Tell,Vid2Vid}, and this paper focuses on the controlling mode of texts.

Early text-to-video generation models~\citep{2018Video_Text,2017Create_What_You_Tell} mainly use convolutional GAN models with recurrent neural networks to model temporal motions. 
Although complex architectures and auxiliary losses are introduced, GAN-based models cannot generate videos beyond simple scenes like moving digits and close-up actions.
Recent works extend text-to-video to open domains with large-scale transformers~\citep{2022MAGVIT} or diffusion models~\citep{2022ImagenVideo}.
Considering the difficulty of high-dimensional video modeling and the scarcity of text-video datasets, training text-to-video generation from scratch is unaffordable.
As a result, most works acquire knowledge from pretrained text-to-image models.
CogVideo~\citep{2023CogVideo} inherits from a pretrained text-to-image model CogView2~\citep{cogview2}.
Imagen Video~\citep{2022ImagenVideo} and Phenaki~\citep{Phenaki} adopt joint image-video training.
Make-A-Video~\citep{MakeAVideo} learns motion on video data alone, eliminating the dependency on text-video data.
To reduce the high cost of video generation, latent diffusion~\citep{2022LDM} has been widely utilized for video generation~\citep{2023LatentShift, VideoLDM, esser2023structure,he2022lvdm, he2022latent, 2023Text2VideoZero, 2023FollowYourPose,wu2022nuwa, wu2022tuneavideo, PVDM, zhou2022magicvideo}.
MagicVideo~\citep{zhou2022magicvideo} inserts a simple adaptor after the 2D convolution layer.
Latent-Shift~\citep{2023LatentShift} adopts a parameter-free temporal shift module to exchange information across different frames.
PDVM~\citep{PVDM} projects the 3D video latent into three 2D image-like latent spaces.
Show-1~\citep{Show_1} combines pixel and latent diffusion.
Although the research on text-to-video generation is very active, existing research ignores the inter and inner correlation between spatial and temporal modules.
In this paper, we revisit the design of text-driven video generation. 

\begin{table*}[t]
    \centering
    \small \renewcommand{\arraystretch}{1.25}
    \caption{Comparison of different open-source datasets with text-video pairs.
    Captions are premium-quality text labels for videos. In contrast, class labels tend to be overly simplistic, and subtitles do not synchronize with the visual contents of the video.
    Of all the open-source datasets available, our HD-VG-130M dataset stands out for its expansive scale, and its labels fulfill the requirements of video generation.
    Furthermore, while many internet videos are unsuitable for training video generation models, most existing datasets fail to adequately filter visual content. Our 40M subset enjoys higher quality (in aspects of visual text, motion, and aesthetics) and offers videos that meet stricter criteria.}
    \vspace{1mm}
    \begin{tabular}{m{4cm}|ccccm{2cm}<{\centering}}
        \Xhline{1.2pt}
        Dataset                            & Video clips & Resolution & Domain & Text & Visual Filtering \\
        \Xhline{0.4pt}
        UCF101~\citeyearpar{soomro2012ucf101}      & 13K         & 240p  & human action & class label & \xmark \\
        ActivityNet 200~\citeyearpar{ActivityNet}                          & 23K         & -     & human action & class label & \xmark \\
        ACAV100M~\citeyearpar{ACAV}                & 100M        & 360p  & open & subtitle &  \xmark \\ 
        HD-VILA-100M~\citeyearpar{xue2022advancing} & 103M       & 720p  & open & subtitle & \xmark  \\ 
        HowTo100M~\citeyearpar{miech2019howto100m} & 136M        & 240p  & instructional & subtitle & \xmark \\ 
        YT-Temporal-180M~\citeyearpar{MERLOT}      & 180M        & -     & open & subtitle & motion \\ 
        \Xhline{0.4pt}
        MSVD~\citeyearpar{MSVD} & 2K & - & open & caption & visual text \\
        YouCook2~\citeyearpar{Youcook2} & 15K & - & cooking & caption & \xmark\\
        MSR-VTT~\citeyearpar{xu2016msr}            &  10K        & 240p  & open & caption & \xmark \\
        VATEX~\citeyearpar{VaTeX} & 41K & - & open & caption & \xmark \\
        LSMDC~\citeyearpar{lsmdc} & 118K & 1080p & movie & caption & \xmark \\
        WebVid-10M~\citeyearpar{bain2021frozen}    & 10M         & 360p  & open & caption & \xmark \\ 
        Panda-70M~\citeyearpar{panda70M}          & 70M         & 720p  & open & caption & \xmark \\
        \Xhline{0.4pt}
        HD-VG-130M (Ours)                    & 130M        & 720p  & open & caption & \xmark \\ 
        \Xhline{0.4pt}
        HD-VG-40M higher-quality subset (Ours)   & 40M         & 720p  & open & caption & visual text, motion, and aesthetics \\ 
        \Xhline{1.2pt}
    \end{tabular}
    \label{tab:hqvg_100m}
\end{table*}

Dataset takes an important role in training text-to-image generative models.
Nonetheless, current datasets either lack the necessary scale or quality~\citep{bain2021frozen}, or are inaccessible to the research community~\citep{SVD}. In this paper, we provide the first open-source high-quality and large-scale dataset.

\section{High-Definition Video Generation Dataset}
\label{sec:hd_vg}

In this section, we construct a large-scale text-video dataset tailored for high-definition, widescreen, and watermark-free video generation. Additionally, We refine the dataset by considering text, motion, and aesthetic factors to create a higher-quality subset.

\subsection{Data Collection, Processing \\ and Annotation}

\begin{figure*}[t]
    \centering
    \includegraphics[width=0.99\linewidth]{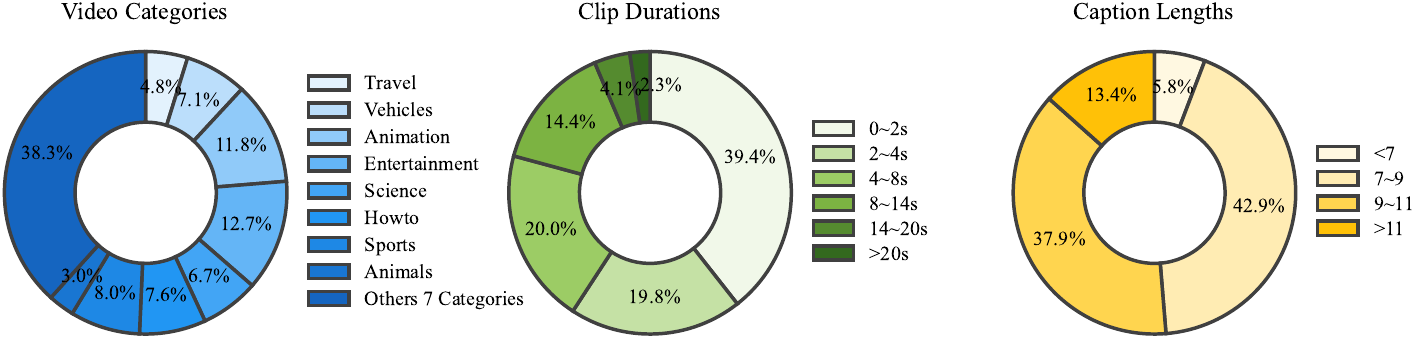}
    \vspace{3mm}
    \caption{Statistics of video categories, clip durations, and caption word lengths in HD-VG-130M. HD-VG-130M covers a wide range of video categories. }
    \label{fig:hdvg_statis}
\end{figure*}

Datasets of diverse text-video pairs are the prerequisite for training open-domain text-to-video generation models.
However, most of existing text-video datasets are limited in either scale or quality, thus hindering the upper bound of high-quality video generation. 
Referring to Table~\ref{tab:hqvg_100m}, MSR-VTT~\citep{xu2016msr} and UCF101~\citep{soomro2012ucf101} only have 10K and 13K video clips respectively.
Although large in scale, HowTo100M~\citep{miech2019howto100m} is specified for instructional videos, which has limited diversity for open-domain generation tasks.
Despite being appropriate in both scale and domain, the formats of textual annotations in HD-VILA-100M~\citep{xue2022advancing} are subtitle transcripts, which lack visual contents related descriptions for high-quality video generation.
Additionally, the videos in HD-VILA-100M have complex scene transitions, which are disadvantageous for models to learn temporal correlations.
WebVid-10M~\citep{bain2021frozen} has been used in some previous video generation works~\citep{2022ImagenVideo, MakeAVideo}, considering its relatively large-scale (10M) and descriptive captions.
Nevertheless, videos in WebVid-10M are of low resolution and have poor visual qualities with watermarks in the center. 

Recently, video generation has attracted considerable attention particularly in the industry, leading to the emergence of several new large-scale text-to-video datasets~\citep{LAVIE,VideoPoet,SVD}.
The LVD~\citep{SVD} dataset provides 577M annotated video clip pairs and demonstrates the importance of large-scale datasets for video generation.
However, as of now, none of these datasets are open source, hindering their use and analysis by other researchers.
Recently released, Panda-70M~\citep{panda70M} is a text-to-video dataset containing 70 million video clips with text annotation. Despite its larger scale compared to existing open-domain datasets, Panda-70M focuses less on data processing, resulting in inappropriate content and limited performance for models trained on it. In Section~\ref{sec:summary_dataset}, more detailed discussions are provided.

To tackle the problems above and achieve high-quality video generation, we propose a large-scale text-video dataset, namely \textbf{HD-VG-130M}, including 130M text-video pairs from open-domain in high-definition (720p), widescreen and watermark-free formats.
We first collect high-definition videos from YouTube.
The \textbf{challenge} lies in converting raw high-definition videos into video-caption pairs, which is far from straightforward.
As the original videos have complex scene transitions which are adverse for models to learn temporal correlations, we detect and split scenes in these original videos,\footnote{We use the open source tool: \url{https://github.com/Breakthrough/PySceneDetect}}
resulting in 130M single scene video clips.
Finally, we caption video clips with BLIP-2~\citep{li2023blip}, in view of its large vision-language pre-training knowledge.
To be specific, we extract the central frame in each clip as the keyframe, and get the annotation for each clip by captioning the keyframe with BLIP-2~\citep{li2023blip}. Note that the video clips in HD-VG-130M are in single scenes, which ensures that the keyframe captions are representative enough to describe the content of the whole clips in most circumstances.
Another method of annotation involves using video captioning techniques. However, we have observed that existing video captioning methods~\citep{mPLUG2} often inaccurately describe the visual content, leading to less effective results compared to BLIP-2. We will delve deeper into this issue in Sec.~\ref{sec:abl_dataset}.

The statistics of HD-VG-130M are shown in Fig.~\ref{fig:hdvg_statis}. The videos in HD-VG-130M cover 15 categories. The wide range of domains is beneficial for training the models to generate diverse content. After scene detection, the video clips are mostly in single scenes with duration less than 20 seconds. The textual annotations are visual contents related to descriptive captions, which are mostly around 10 words.

\begin{figure*}[t]
    \centering
    \includegraphics[width=\linewidth]{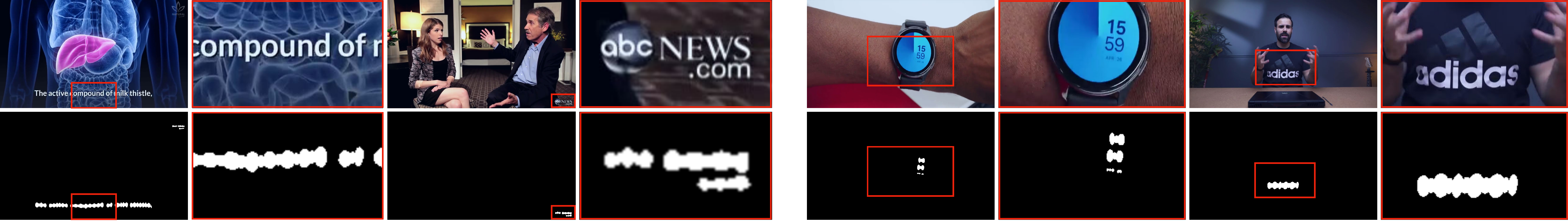}
    \begin{minipage}{0.48\linewidth}
        \subcaption{Videos filtered out}
    \end{minipage}
    \begin{minipage}{0.05\linewidth}
    \end{minipage}
    \begin{minipage}{0.48\linewidth}
        \centering
        \subcaption{Video not filtered out}
    \end{minipage}
    \caption{Results of our filtering strategy on visual texts. The first row shows the frame contents, and the second row shows the predictions of the text detector. On the right, videos containing text but no channel names in the corner nor subtitles are not filtered out, supporting the diversity of our dataset.}
    \label{fig:logo_case}
\end{figure*}

\subsection{Further Data Processing}

Despite detecting and splitting scenes, numerous videos remain unsuitable for training high-quality text-to-video generation models. 
Given that the videos are sourced from YouTube, a subset of them displays YouTube channel names or subtitles.
Additionally, some videos are entirely static or consist of images with simple transformation animations.
Such videos negatively impact the training of text-to-video generation models.
However, existing text-video datasets overlook the significance of filtering visual content. MSVD~\citep{MSVD} manually removes videos containing subtitles or overlaid text, but manual processing is impractical for handling large-scale data. YT-Temporal-180M~\citep{MERLOT} adopts a basic strategy to remove static videos based on their four thumbnails, which is of low precision. Additionally, aesthetic quality is rarely taken into account. HowTo100M~\citep{miech2019howto100m} and HD-VILA-100M~\citep{xue2022advancing} employ a simplistic approach by retaining videos with high view counts; however, a high view count does not guarantee video quality. In the following section, we provide a detailed discussion on visual filtering and propose methods to address these issues and create a higher-quality subset.

\begin{figure*}[t]
    \centering
    \includegraphics[width=0.99\linewidth]{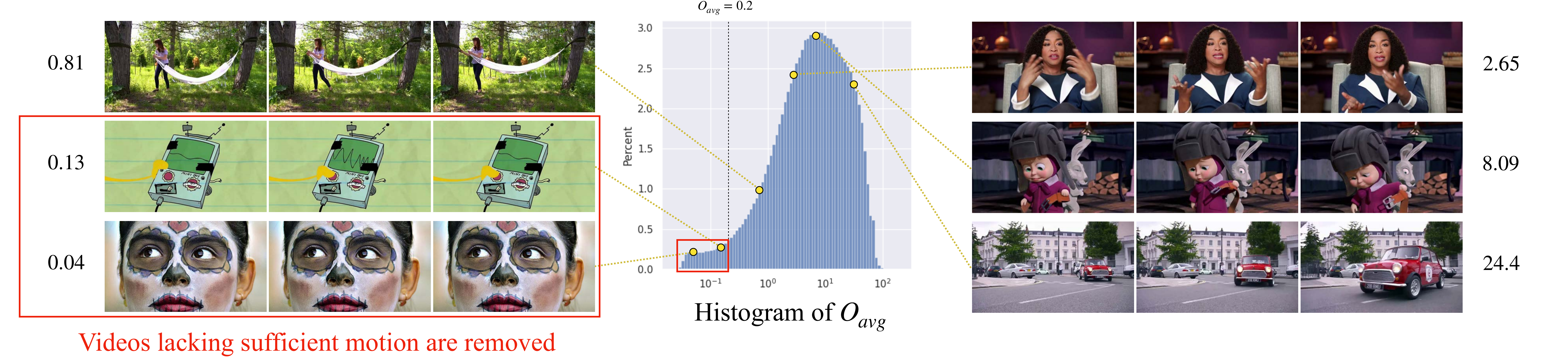}
    \vspace{3mm}
    \caption{The distribution of the average optical flow magnitude $O_{avg}$ and representative samples across different $O_{avg}$ values. Videos with $O_{avg}<0.2$ demonstrate minimal motion, which is unsuitable for training text-to-video models effectively. Hence, we exclude these videos and retain only those with $O_{avg}>0.2$, which indicate significant motion.}
    \label{fig:motion_case}
\end{figure*}

\begin{figure*}[t]
    \centering
    \includegraphics[width=\linewidth]{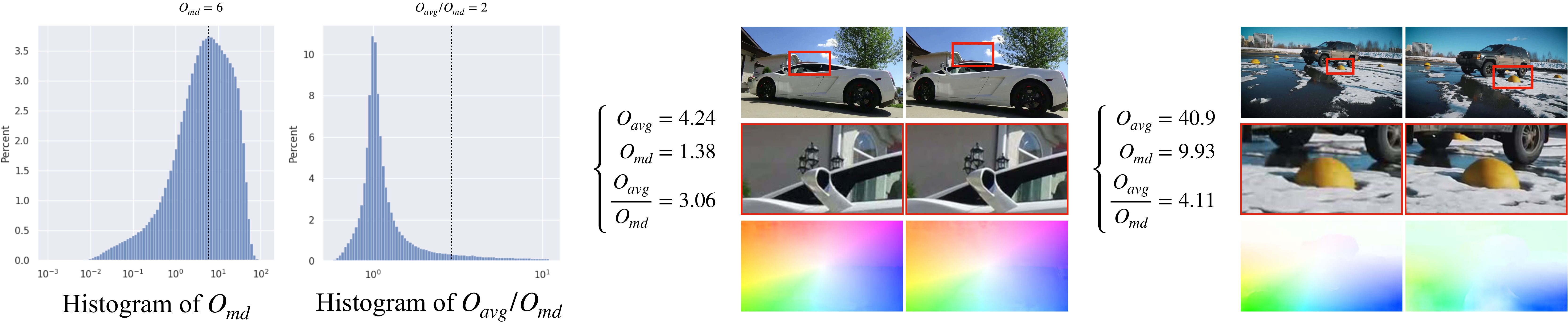}
    \begin{minipage}{0.4\linewidth}
        \subcaption{Distributions of $O_{md}$ and $O_{avg}/O_{md}$}
    \end{minipage}
    \begin{minipage}{0.325\linewidth}
        \centering
        \subcaption{Image zooming transformation}
    \end{minipage}
    \begin{minipage}{0.26\linewidth}
        \centering
        \subcaption{Real-world zooming}
    \end{minipage}
    \caption{The distribution of the mean deviation of optical flow $O_{md}$, and the ratio of $O_{avg}/O_{md}$. Videos with high $O_{avg}/O_{md}$ tend to exhibit consistent optical flows overall, often indicating global translation or scaling. Additionally, among these videos, real-world scenes typically have higher $O_{md}$ values. We show two samples for static image transformation in (b) and real-world camera transformation in (c). In (b), the zoomed details indicate that the scene is static, while in (c), the zoomed details show that the relative position of objects has changed. We also show optical flows for better illustration.}
    \label{fig:motion_diversity_case}
\end{figure*}

\subsubsection{Text Detection}
As illustrated in Fig.~\ref{fig:logo_case}(a), since videos are collected from YouTube, some of them contain channel names in the corners or subtitles at the bottom half. These videos may lead the text-to-video model to generate texts that have nothing to do with the video content, which goes against the intended purpose of users.

We utilize optical character recognition to identify and filter these videos. Specifically, we employ the text detector CRAFT~\citep{CRAFT} to locate textual elements. Note that while we want to remove channel names and subtitles, we do not want to remove all videos that contain text, which would reduce the diversity of the dataset. 
As shown in Fig.~\ref{fig:logo_case}(b), text can be found on various goods and clothing items, which are quite common in the real world.
Therefore, we only consider text within the $H_{text}$ pixel range from the upper, lower, left, and right edges.
The keyframes selected are identical to those employed in the aforementioned captioning process. Considering speed and precision, videos are resized to 640 pixels width and $H_{text}$ is set to 60.
This strategy results in the removal of 37.33\% of videos. Among the remaining videos, 73.36\% still contain text, which supports the diversity of our dataset.

\subsubsection{Motion Detection}
\label{sec:motion_detection}

We employ the PWC-Net optical flow estimator \citep{PWCNet} to analyze video motion. To minimize computational demands, videos are sampled at a rate of 2 frames per second (FPS). Two scores are computed: the average optical flow magnitude ($O_{avg}$) and the mean deviation of optical flows ($O_{md}$). Videos shorter than 2 seconds lack sufficient frames for extraction at 2 FPS, so we exclude them when constructing the higher-quality subset.

Generally, the distribution of real-world optical flow magnitude $O_{avg}$, should be similar to the Gaussian distribution. However, as shown in Fig.~\ref{fig:motion_case}, the area within the red box does not conform to the tail shape of a Gaussian distribution. This is because internet videos may contain scenes that depict an image, and corresponding video clips may remain completely still. These cases of insufficient motion could mislead the video generative model. To eliminate these instances, we apply a filtering rule of $O_{avg} > 0.2$, resulting in the removal of 3.71\% of videos.

Some scenes may not be static, but rather consist solely of an image with translation or scaling transformations. An example is shown in Fig.~\ref{fig:motion_diversity_case}(b), where an image of a car is slowly zoomed in. These movements are overly simplistic and fail to accurately reflect real-world object motion, thereby diminishing the effectiveness of video generative models. Such global transformations can be readily identified using the ratio of $O_{avg}/O_{md}$. $O_{md}$ signifies the diversity across frames in the optical flow. When the value of $O_{md}$ is lower than $O_{avg}$, it indicates that the motion across frames is largely uniform, typically indicative of global transformations such as translation and scaling.

One issue with this filtering strategy is that video clips involving camera zooming, scaling, and translation also demonstrate high $O_{avg}/O_{md}$ values. We observe that in such real-life scenarios, changes in the viewing angle can cause shifts in object occlusion relationships. This is illustrated in Fig.~\ref{fig:motion_diversity_case}(c), where the background surrounding the yellow ball undergoes a change. These variations in content lead to inconsistent optical flow across frames, resulting in relatively high $O_{md}$ values.
Therefore, we finally employ a filtering strategy in which we keep videos satisfying either $O_{avg}/O_{md}<2$ or $O_{md}>6$, which is able to remove image transformation animations while retaining real-world camera transformations. It removes 9.58\% of videos from the dataset.

\begin{figure*}[t]
    \centering
    \includegraphics[width=0.98\linewidth]{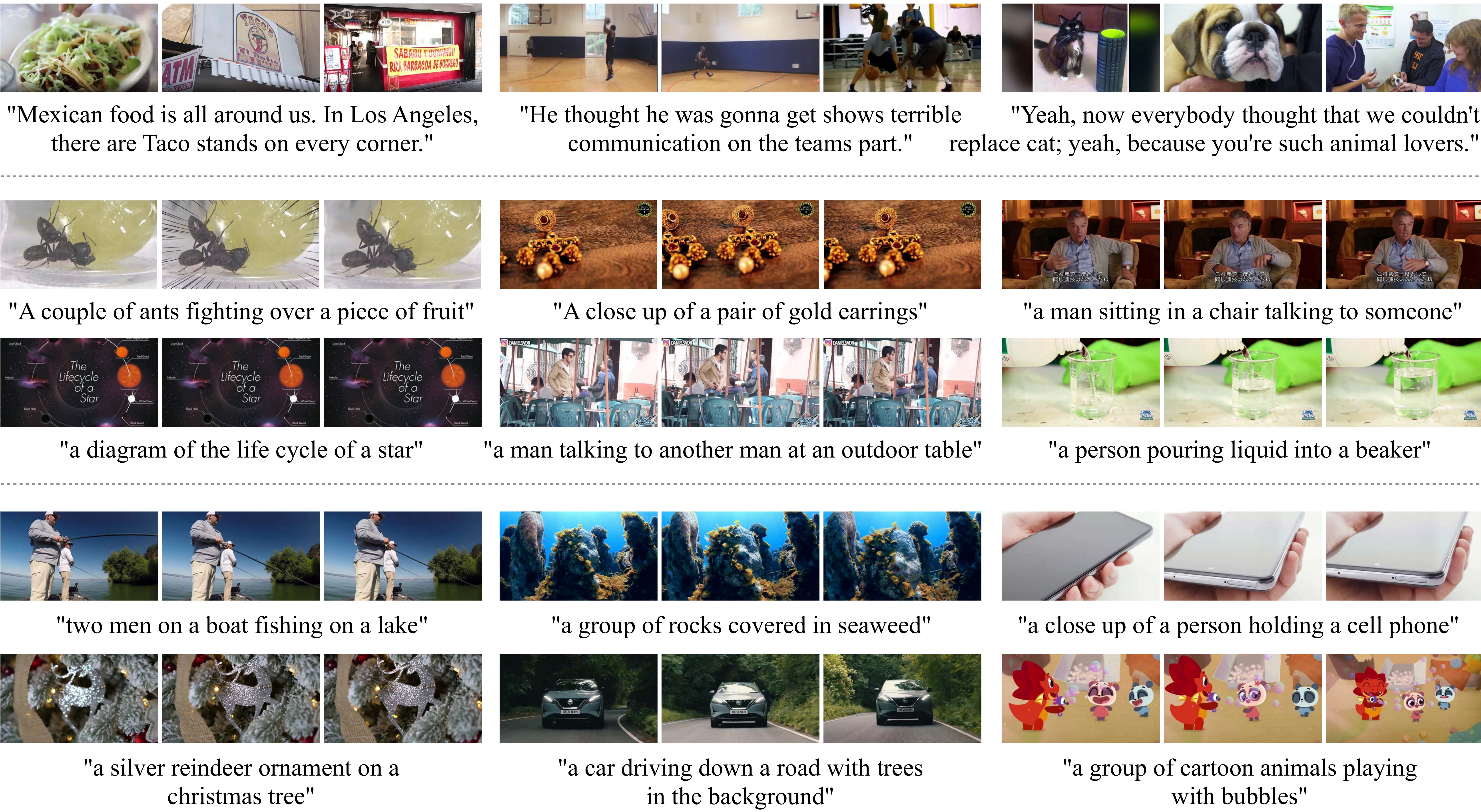}
    \vspace{2mm}
    \caption{Video-caption pairs in various datasets. From top to bottom: HD-VILA-100M, HD-VG-130M (\textit{excluding} HD-VG-40M), and HD-VG-40M. Compared to HD-VG-130M, HD-VILA-100M videos lack coherence in both visuals and accompanying text. In HD-VG-40M, static scenes and meaningless text are filtered out, enhancing the dataset's quality for text-to-video generation.}
    \label{fig:hdvg_preview}
\end{figure*}

\begin{figure}[t]
    \centering
    \includegraphics[width=0.98\linewidth]{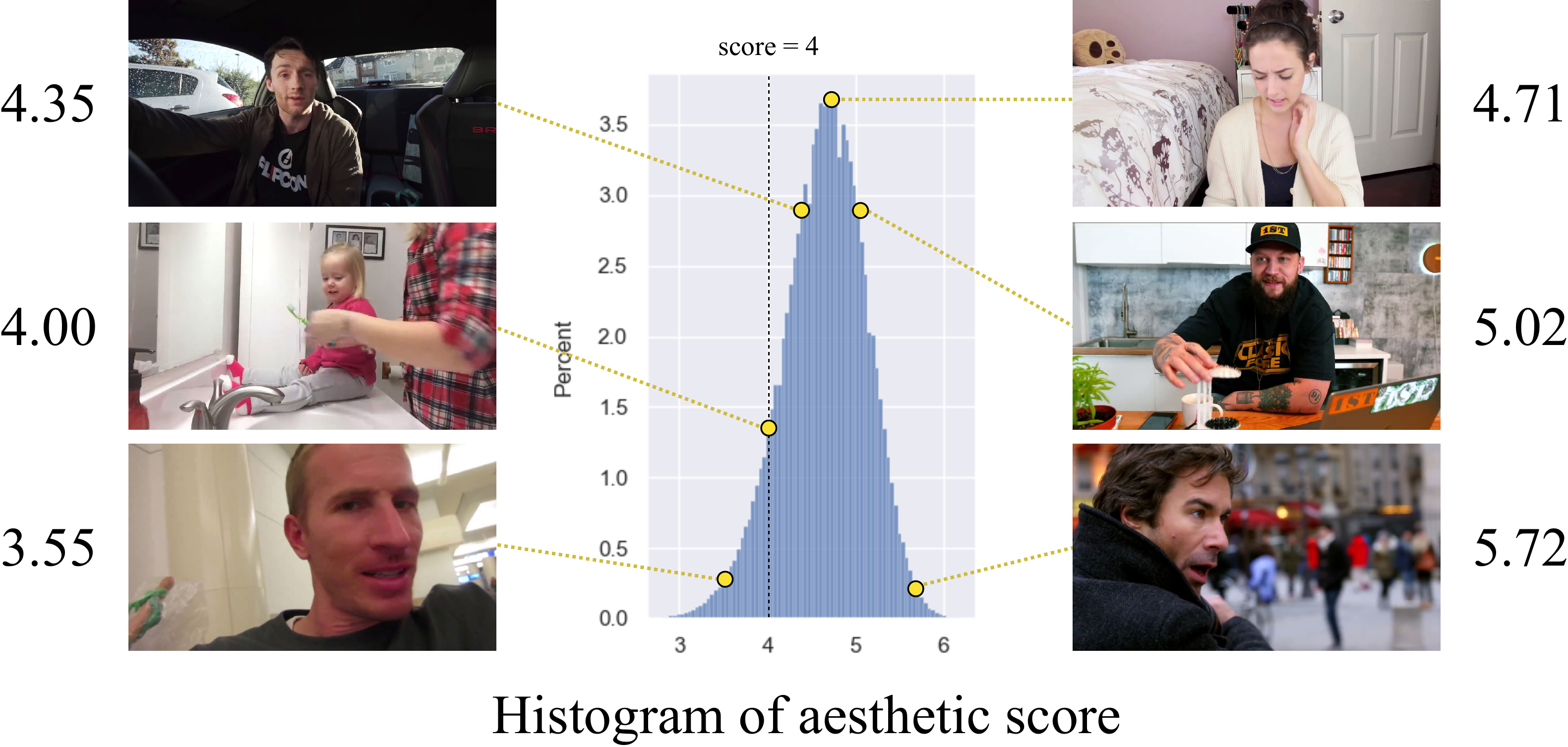}
    \vspace{2mm}
    \caption{The distribution of aesthetic scores alongside samples depicting the same theme (human) across various aesthetic scores. Videos with scores above 4 exhibit good aesthetic quality.}
    \label{fig:aesthetics}
\end{figure}

\begin{figure}[t]
    \centering
    \includegraphics[width=0.98\linewidth]{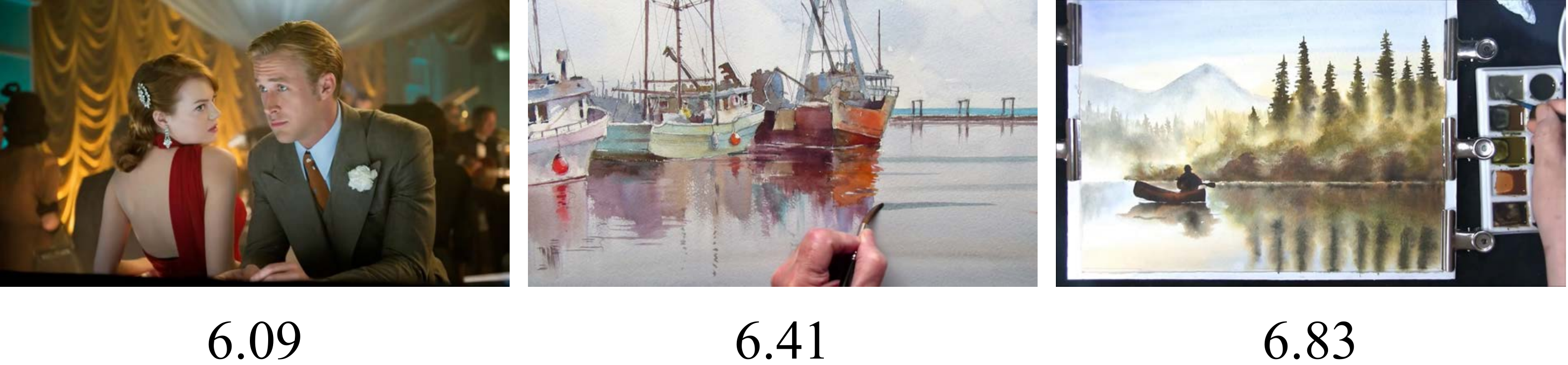}
    \vspace{2mm}
    \caption{Samples from our dataset with aesthetic scores exceeding 6. Art films typically score around 6, while videos with scores exceeding 6.5 mostly feature people drawing.}
    \label{fig:high_aesthetics}
\end{figure}

\subsubsection{Aesthetics Evaluation}

We apply the widely-used LAION-Aesthetics Predictor V2\footnote{https://github.com/christophschuhmann/improved-aesthetic-predictor} to evaluate the aesthetics of video frames.
The distribution of aesthetic scores is shown in Fig.~\ref{fig:aesthetics}.
We also provide samples of humans to compare aesthetic effects within the same theme.
Videos with aesthetic scores of 4 and below are usually uploaded by ordinary users. Although the content is clear, the composition and lighting are relatively random, and the contrast is low.
Videos with an aesthetic score around 4.7, \ie, the majority of the dataset, have standard composition and aesthetic effects in line with mainstream aesthetics.
Videos with an aesthetic score closer to 6 have more artistic effects, such as asymmetrical composition or exaggerated background blurring.
To enhance the beauty of the data, we filtered out the samples with an aesthetic score below 4 and removed 9.37\% of the videos.

Very high-quality videos, such as art film slices, typically have an aesthetic score of around 6.
Few samples have aesthetic scores of 6.5 and above.
This is because the LAION-Aesthetics Predictor V2 tends to give higher scores to artistic paintings rather than realistic scenes.
For videos with aesthetic scores higher than 6.5, many of them are about static painting images, which are removed by our motion filtering in Sec.~\ref{sec:motion_detection}.
The remaining videos mostly depict people painting, as shown in the right two samples in Fig.~\ref{fig:high_aesthetics}.

\subsubsection{Summary}
\label{sec:summary_dataset}

After implementing the aforementioned data processing steps, the HD-VG-130M dataset is refined into a higher-quality subset of 40 million samples, addressing issues such as meaningless texts, lack of movement, and low aesthetics. This subset is named HD-VG-40M.
A visualization comparing data samples from HD-VILA-100M, HD-VG-130M, and HD-VG-40M is presented in Fig.~\ref{fig:hdvg_preview}.
In comparison to HD-VG-130M, the videos in HD-VILA-100M lack semantic coherence, and the accompanying text fails to describe the video contents.
In HD-VG-40M, videos containing static scenes and meaningless text are further filtered out, resulting in higher-quality data for text-to-video generation.
Despite removing more than half of the samples, our higher-quality subset remains larger than most of the existing open-source text-to-video generation datasets, as shown in Table~\ref{tab:hqvg_100m}.
We will demonstrate later that fine-tuning with our higher-quality subset can further enhance the performance of video generation.

\begin{figure}[t]
    \centering
    \includegraphics[width=0.98\linewidth]{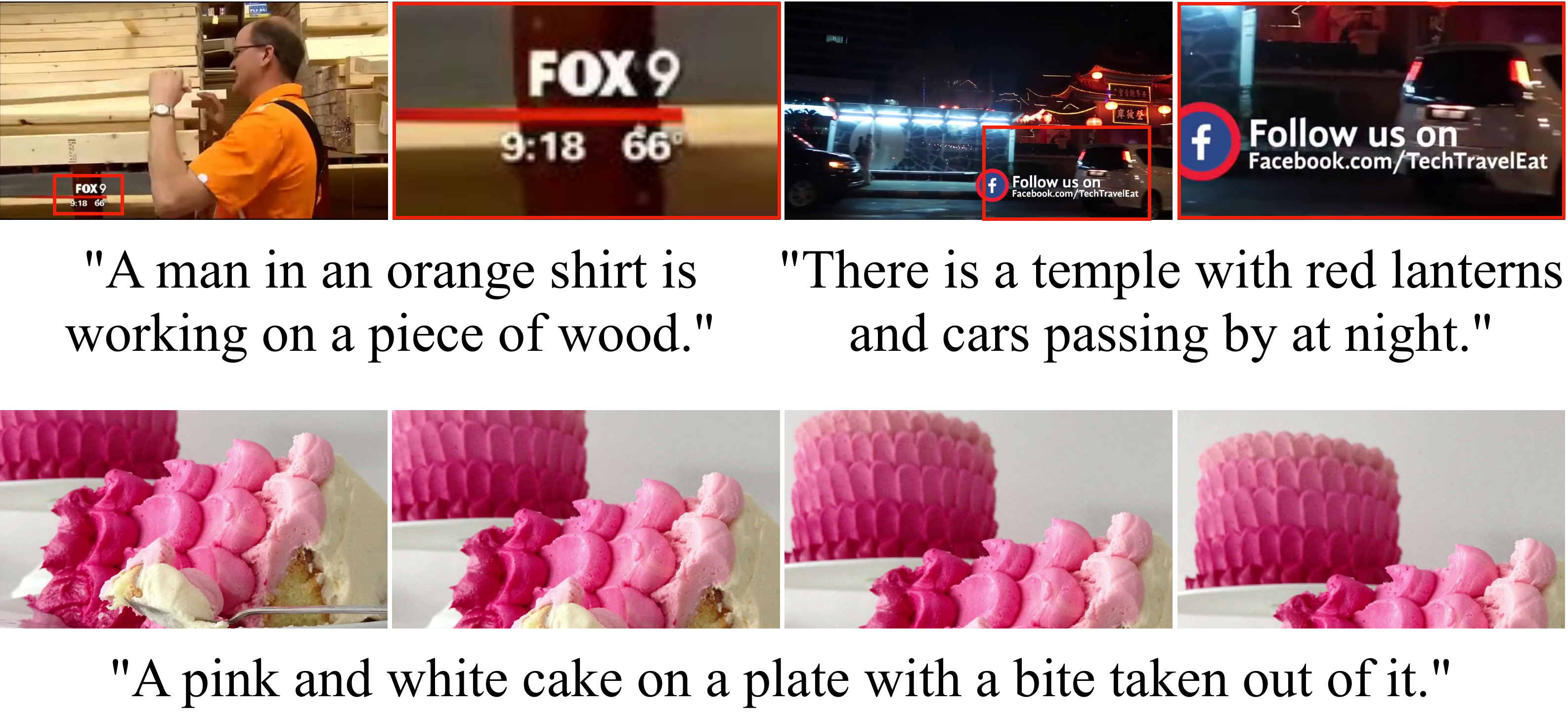}
    \vspace{2mm}
    \caption{The Panda-70M dataset contains video samples where the text is independent of the picture content, as well as videos consisting solely of the translation of static images.}
    \label{fig:failure_panda70m}
\end{figure}

Panda-70M~\citep{panda70M} is a recently released large-scale text-video dataset, making a significant contribution to the AIGC community.
Panda-70M focuses more on meticulous video captioning but somehow neglects the importance of visual content filtering.
Both our dataset and Panda-70M collect data from YouTube. However, as discussed above, not all internet videos are suitable for training video generation models, leading to improper samples in Panda-70M, as illustrated in Fig.~\ref{fig:failure_panda70m}.
In comparison, our work conducts detailed processing on the visual contents, filling the gaps left by open-source data in this area.
Moreover, we introduce a novel spatiotemporal interaction strategy to enhance model design.
Consequently, our model exhibits enhanced visual quality and text-video alignment compared to the model trained on Panda-70M in Tables~\ref{tab:result_ucf101}-\ref{tab:result_msrvtt}.

\section{High-Quality Text-to-Video Generation}
\label{sec:method}

In this section, we introduce how we build the text-to-video generation framework. We first describe how we reinforce both spatial and temporal interactions. Then, we introduce the detailed architecture of our model and the super-resolution processing for generating high-definition videos.

\subsection{Spatiotemporal Connection}
To reduce computational costs and leverage pretrained image generation models, space-time separable architectures have gained popularity in text-to-video generation~\citep{2022_VDM,2023CogVideo}.
These architectures handle spatial operations independently on each frame, while temporal operations consider multiple frames for each spatial position. 
In the following, we refer to the features predicted by 2D/spatial modules in space-time separable networks as ``spatial features", and ``temporal features'' vice versa.

\begin{figure*}[t]
    \centering
    \includegraphics[width=0.99\linewidth]{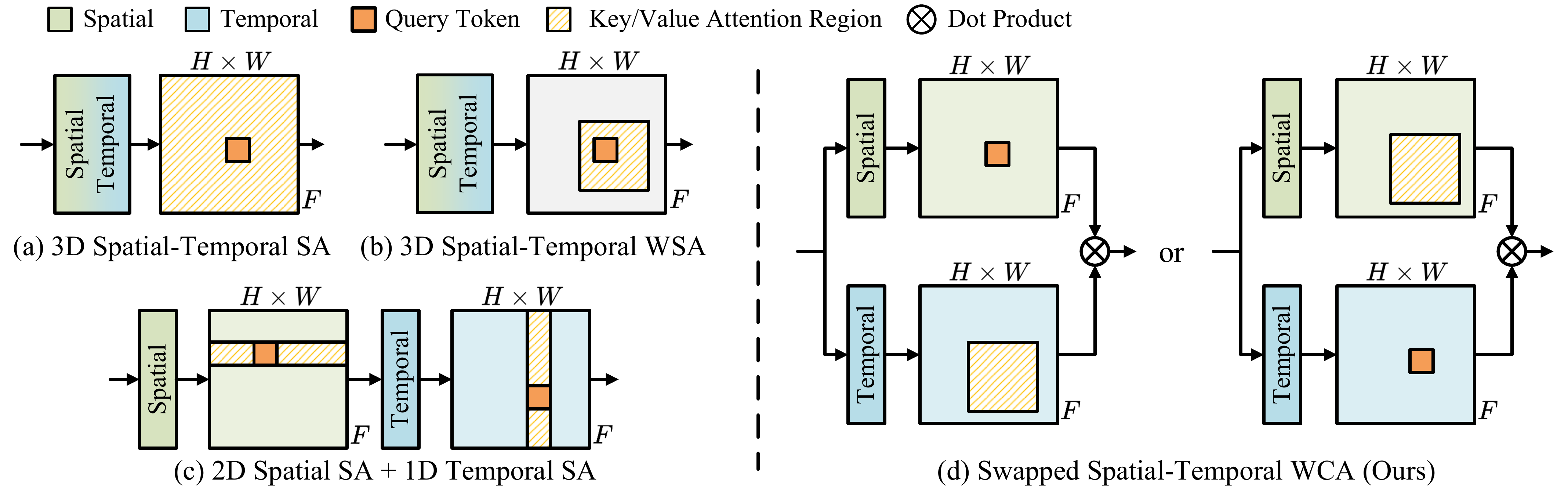}
    \vspace{3mm}
    \caption{The paradigm of {the proposed} Swapped spatiotemporal Cross-Attention (Swap-CA) in comparison with existing video attention schemes. Instead of only conducting self-attention in (a)-(c), we perform cross-attention between spatial and temporal modules in a U-Net, which encourages more spatiotemporal mutual reinforcement.}
    \label{fig:teaser_attn}
\end{figure*}

The quality of spatiotemporal features is important for video generation, as it can affect temporal consistency and text-content alignment performance~\citep{2023CogVideo,2022ImagenVideo}.
The interaction between spatial and temporal features is also essentially, as it determines how the spatial and temporal features are combined.
This interaction has been highlighted in previous video-related studies~\citep{2021_space_time_attn,ZengFC20} and verified in cross-modality learning~\citep{abs-2304-11335,MMDiffusion}.
However, as discussed in Sec.~\ref{sec:introduction}, prior works have neglected the crucial interaction between spatial and temporal features. The methodologies of existing spatiotemporal strategies are illustrated in Fig.~\ref{fig:teaser_attn}~(a)-(c). None of them capture the interaction between spatial and temporal features. To address this limitation, we propose the mutual reinforcement of these features through a series of cross-attention operations. As shown in Fig.~\ref{fig:teaser_attn}~(d), our swap attention mechanism enhances bidirectional guidance between spatial and temporal features by treating one feature as the query and the other as the key/value. To ensure the reciprocity of information flow, we also interchange the role of the ``query" in adjacent layers. In the following, we introduce the details of this design.

First, denote a basic operation
\begin{equation}
    \text{CrossAttention}(x,y)=\text{softmax}(\frac {QK^{T}} {\sqrt{d}})\cdot V,
\end{equation}
with
\begin{align}
    Q = W^{(i)}_Q \cdot x, \;\ 
    K = W^{(i)}_K \cdot y, \;\ 
    V = W^{(i)}_V \cdot y,
\label{eq:cross_attn}
\end{align}
where $W^{(i)}_Q$, $W^{(i)}_K$, and $W^{(i)}_V$ are learnable projection matrices in the $i$-th layer.
The direction of cross-attention, specifically whether $Q$ originates from spatial or temporal features, plays a decisive role in determining the impact of cross-attention.
In general, spatial features tend to encompass a greater amount of contextual information, which can improve the alignment of temporal features with the input text. On the other hand, temporal features have a complete receptive field of the time series, which may enable spatial features to generate visual content more effectively.
To leverage both aspects effectively, we propose a strategy of swapping the roles of $Q$ and $K,V$ in adjacent two blocks. This approach ensures that both temporal and spatial features receive sufficient information from the other modality, enabling a comprehensive and mutually beneficial interaction.

\begin{figure*}[t]
    \centering
    \includegraphics[width=0.99\linewidth]{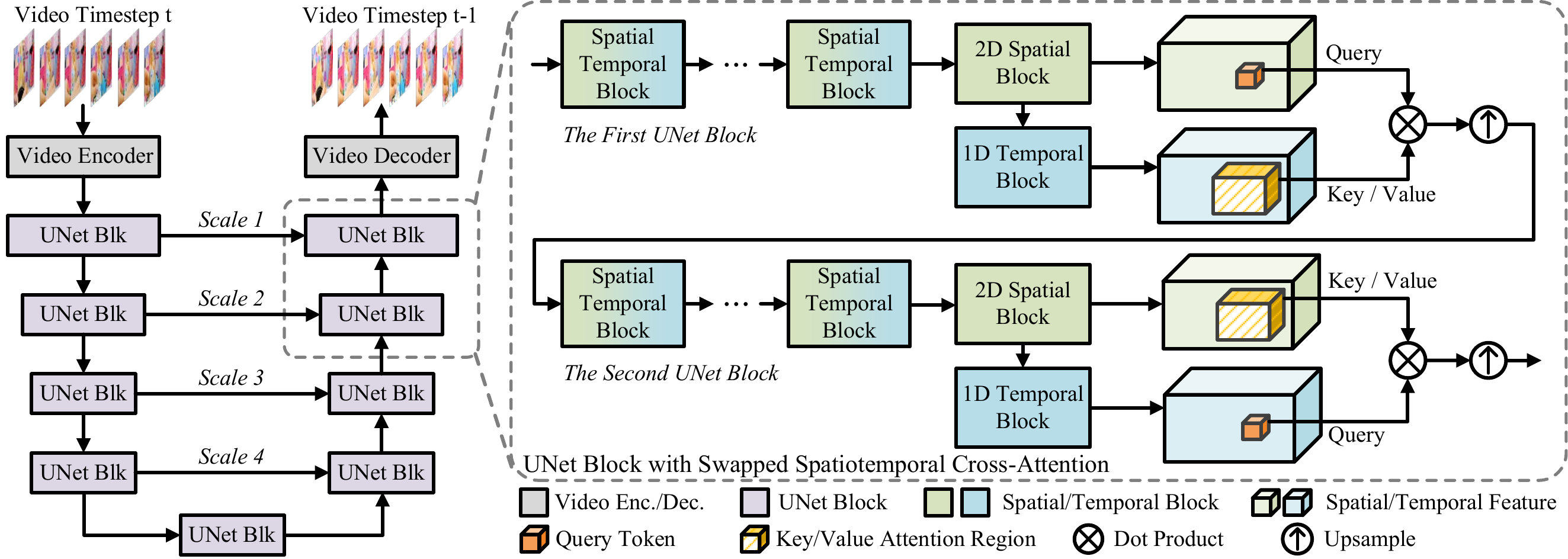}
    \vspace{3mm}
    \caption{An illustration of our video diffusion model incorporating Swapped spatiotemporal Cross-Attention (Swap-CA). At the end of each U-Net block, we employ a swapped cross-attention scheme on 3D windows to facilitate a comprehensive integration of spatial and temporal features. In the case of two consecutive blocks, the first block employs temporal features to guide spatial features, while in the second block, their roles are reversed. This reciprocal arrangement ensures a balanced and mutually beneficial interaction between the {spatiotemporal} modalities throughout the model.}
    \label{fig:win_attn}
\end{figure*}

Global attention greatly increases the computational costs in terms of memory and running time. 
To improve efficiency, we conduct 3D window attention.
Given a video feature in the shape of $F\times H \times W$ and a 3D window size of $F_w \times H_w \times W_w$, we organize the windows to process the feature in a non-overlapping manner, leading to $\lceil \frac{F}{F_w} \rceil \times \lceil \frac{H}{H_w} \rceil \times \lceil \frac{W}{W_w} \rceil$ distinct 3D windows.
Within each window, we perform spatiotemporal cross-attention. By adopting the 3D window scheme, we effectively reduce computational costs without compromising performance.

Following prior text-to-image arts~\citep{VideoLDM, 2022LDM}, we incorporate 2$\times$ down/upsampling along the spatial dimension to establish a hierarchical structure.
Furthermore, research~\citep{2019VideoCompression,2020VideoCompression} has pointed out that the temporal dimension is sensitive to compression.
In light of these considerations, we do compress the temporal dimension and conduct shift windows~\citep{2022VideoSwin}, which advocates an inductive bias of locality.
On the spatial dimension, we do not shift since the down/upsampling already introduces connections between neighboring non-overlapping 3D windows.

To this end, we propose a Swapped spatiotemporal Cross-Attention (Swap-CA) in 3D windows.
Let $t^l$ and $s^l$ represent the predictions of 2D and 1D modules. We utilize Multi-head Cross Attention (MCA) to compute their interactions by Swap-CA as
\begin{equation}
\begin{aligned}
    \Tilde{s}^l & = \text{Proj}^l_{in} \odot \text{GN}(s^l); \\
    \Tilde{t}^l &= \text{Proj}^l_{in}\odot \text{GN}(t^l); \\
    h^l &= \text{3DW-MCA}(\text{LN}(\Tilde{s}^l), \; \text{LN}(\Tilde{t}^l)) + \Tilde{s}^l; \\
    \bar{h}^l &= \text{FFN} \odot \text{LN} (h^l) + h^l; \\
    z^l & = t^l + s^l + \text{Swap-CA}(s^l, t^l) \\
        & = t^l + s^l + \text{Proj}^l_{out}(\bar{h}^l),
\end{aligned}
\label{eq:first_layer}
\end{equation}
where {Group Norm} (GN), {Projection} (Proj), {Layer Norm} (LN), and 3D Window-based Multi-head Cross-Attention (3DW-MCA) are learnable modules.
By initializing the output projection Proj$^{l-1}_{out}$ by zero, we have $z^l = t^{l-1} + s^{l-1}$, \ie, Swap-CA is skipped so that it is reduced to a basic addition operation.
This allows us to initially train the diffusion model using addition operations, significantly speeding up the training process.
Subsequently, we can switch to Swap-CA to enhance the model's performance.

Then for the next spatial-temporal separable block, we apply 3D Shifted Window Multi-head Cross-Attention (3DSW-MCA) and interchange the roles of $s$ and $t$, as
\begin{equation}
    h^{l+1} = \text{3DSW-MCA}(\text{LN}(\Tilde{t}^{l+1}), \text{LN}(\Tilde{s}^{l+1})) + \Tilde{t}^{l+1}.
\label{eq:second_layer}
\end{equation}
In all 3DSW-MCA, we shift the window along the temporal dimension by $\lceil \frac{F_w} 2 \rceil$ elements.

\subsection{Overall Architecture}
We adopt the LDM~\citep{2022LDM} model as the text-to-image backbone.
We employ an auto-encoder to compress the video into a down-sampled 3D latent space.
Within this latent space, we perform diffusion optimization using an hourglass spatial-temporal separable U-Net model.
Text features are extracted with a pretrained CLIP~\citep{CLIP} model and inserted into the U-Net model through cross-attention on the spatial dimension.

\begin{table*}[t]
    \centering
    \small \renewcommand{\arraystretch}{1.25}
    \caption{Ablation study on spatiotemporal interaction strategies. We report the FVD~\citep{FVD} and CLIPSIM~\citep{CLIP} on 1K samples from the WebVid-10M~\citep{bain2021frozen} validation set. Computational cost is evaluated on inputs of shape $4\times16\times32\times32$. Details can be found in the \textit{appendix}. $T$ and $S$ represent spatial and temporal features, respectively.}
    \vspace{1mm}
    \begin{tabular}{c|c|c|ccccc}
        \Xhline{1.2pt}
        Attention Type   & $Q$ & $K,V$ & Param. (G) & Mem. (GB) & Time (ms) &  FVD $\downarrow$ & CLIPSIM $\uparrow$ \\
        \Xhline{0.4pt}
        -                  & -    &   -  & 1.480 & 9.37  & 135.35 & 566.16  & 0.3070 \\ 
        \Xhline{0.4pt}
                            & $T$ & $S$  & 1.601 & 22.96 & 202.12 & 555.35 & 0.3091 \\ 
        Global              & $S$ & $T$  & 1.601 & 22.96 & 205.00 & 496.25 & 0.3073 \\
        \cmidrule{2-8}
                           & \multicolumn{2}{c|}{Swapped}  & 1.601 & 22.96 & 201.51 & 485.86 & 0.3092 \\
        \Xhline{0.4pt}
                           & $T$ & $S$  & 1.601 & 9.83 & 150.49 & 563.12 & 0.3086 \\ 
        3D Window           & $S$ & $T$   & 1.601 & 9.83 & 149.93 & 490.60 & 0.3076 \\
        \cmidrule{2-8}
                            & \multicolumn{2}{c|}{Swapped}  & 1.601 & 9.83 & 148.24 & \textbf{475.09} & \textbf{0.3107} \\
        \Xhline{1.2pt}
    \end{tabular}
    \label{tab:aba_interaction}
\end{table*}

The detailed architecture of our framework is illustrated in Fig.~\ref{fig:win_attn}.
To balance performance and efficiency, we use Swap-CA only at the end of each U-Net encoder and decoder block. In other positions, we employ a straightforward fusion technique using a $1$$\times1$$\times1$ convolution to merge spatial and temporal features.
To enhance the connectivity among temporal modules, we introduce skip connections that connect temporal modules separated by spatial down/upsampling modules.
This strategy promotes stronger integration and information flow within the temporal dimension of the network architecture.

\subsection{Super-Resolution Towards Higher Quality}
To obtain visually satisfying results, we further perform Super-Resolution (SR) on the generated video. One key to improving SR performance is designing a degradation model that closely resembles the actual degradation process~\citep{Wang_2021_RealESRGAN_ICCV}. In our scenario, the generated video quality suffers from both the diffusion and auto-encoder processes. Therefore, we adopt the hybrid degradation model in Real-ESRGAN~\citep{Wang_2021_RealESRGAN_ICCV} to simulate possible quality degradation caused by the generated process. 
During training, an original video frame is downsampled and degraded using our model, and the SR network attempts to perform SR on the resulting low-resolution image. 
We adopt RCAN~\citep{zhang2018_RCAN} with 8 residual blocks as our SR network. It is trained with a vanilla GAN~\citep{GAN} to improve visual satisfaction. With a suitable degradation design, our SR network can further reduce possible artifacts and distortion in the frames, increase their resolution, and improve their visual quality.

\section{Experiments}
\label{sec:experiments}

In this section, we present the experimental results on text-to-video generation. We first introduce the implementation details, then provide an analysis of method design, and finally compare the performance with existing methods.

\subsection{Implementation Details}

Our model predicts images at a resolution of 344$\times$192 (with a latent space resolution of 43$\times$24). Then a 4$\times$upscaling is produced in our SR model, resulting in a final output resolution of $1376\times768$.
Our model is trained with 32 NVIDIA V100 GPUs.
We utilize our HD-VG-130M as training data to promote the generation visual qualities. 
Furthermore, considering that the textual captions in HD-VG-130M are annotated by BLIP-2~\citep{li2023blip}, which may have some discrepancies with human expressions, we adopt a joint training strategy with WebVid-10M~\citep{bain2021frozen} to ensure the model could generalize well to diverse humanity textual inputs. 
This approach allows us to benefit from the large-scale text-video pairs and the superior visual qualities of HD-VG-130M while maintaining the generalization ability to diverse textual inputs in real scenarios, enhancing the overall training process.
Our model is finally fine-tuned on the HD-VG-40M subset to further promote the performance.
More details can be found in the \textit{appendix}.

\subsection{Ablation Studies}
\label{sec:ablation}

{In this section, we conduct in-depth analyses of the designs of our text-to-video generation model and the construction of our dataset.}

\begin{figure*}[t]
    \centering
    \includegraphics[width=0.99\linewidth]{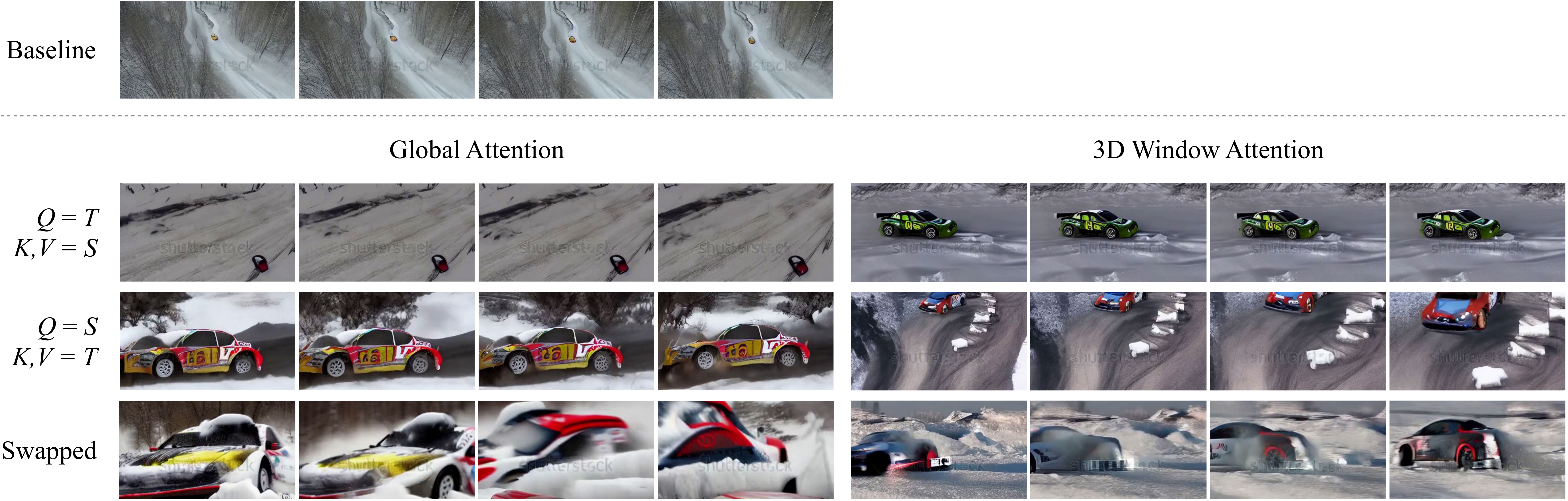}
    \vspace{3mm}
    \caption{Subjective ablation study results. The input prompt is ``Rally racing car ice racing, realistic''.}
    \label{fig:ablation_subjective}
\end{figure*}

\subsubsection{Spatiotemporal Inter-Connection}
We first evaluate the design of our swapped cross-attention mechanism. 
As shown in Table~\ref{tab:aba_interaction}, using temporal features as $Q$ generally leads to better CLIP similarity (CLIPSIM)~\citep{CLIP}, revealing a better text-video alignment.
The reason might be that language cross-attention only exists in spatial modules. Thus, using spatial features to guide temporal ones implicitly enhance semantic guidance.
Reversely, using spatial as $Q$ leads to significantly better FVD, revealing better video quality. The reason might be that the spatial features can better perceive the overall video by using temporal features as guidance.
This experiment demonstrates the benefits of introducing cross-attention, as well as the different acts of spatial and temporal features.
Combining these two aspects, we propose to swap the roles of $x$ and $y$ every two blocks.
In this way, both the temporal and spatial features can get sufficient information from the other modality, leading to improved FVD and CLIPSIM scores.
3D window attention not only significantly lowers computational costs but also leads to a slight performance improvement. Previous studies~\citep{LocalViT,KVT} have observed similar performance improvements by integrating a module to enhance local information within transformer-like structures. 
We show comparative examples in Fig.~\ref{fig:ablation_subjective}.
These results demonstrate how our cross-attention design distinctly enhances scene quality and video dynamics.

\begin{table}
    \centering
    \small \renewcommand{\arraystretch}{1.25}
    \caption{Ablation study on attention strategies.}
    \vspace{1mm}
    \begin{tabular}{l|cc}
        \Xhline{1.2pt}
        Methods & FVD $\downarrow$ & CLIPSIM $\uparrow$ \\
        \Xhline{0.4pt}
        Baseline & 566.16 & 0.3070 \\
        \Xhline{0.4pt}
        Tune-A-Video~\citeyearpar{wu2022tuneavideo} & 717.34	& 0.3084 \\
        CogVideo~\citeyearpar{2023CogVideo} & 534.48	& 0.3010 \\
        3D Spatiotemporal WSA & 500.49	& 0.3072 \\ 
        \Xhline{0.4pt}
        Swap-CA (Ours) & \textbf{475.09} & \textbf{0.3107} \\
        \Xhline{1.2pt}
    \end{tabular}
    \label{tab:aba_attention}
\end{table}

We conduct comparisons with other attention strategies in Table~\ref{tab:aba_attention}. We re-implement these designs within our framework.
Specifically, 3D spatial-temporal WSA is realized by first adding spatial and temporal features together and then applying 3D window self-attention.
All other settings remain consistent with the setting in Table~\ref{tab:aba_interaction}.
The custom attention mechanism utilized in the one-shot model, Tune-A-Video~\citep{wu2022tuneavideo}, appears to be less effective in the open-domain setting.
While CogVideo~\citep{2023CogVideo} and 3D spatial-temporal WSA surpass the baseline, they bring less performance improvement compared with our Swap-CA, showing the effectiveness of our spatiotemporal interaction approach.

\begin{table*}[t]
    \centering
    \small \renewcommand{\arraystretch}{1.25}
    \caption{Ablation study on attention window size.}
    \vspace{1mm}
    \begin{tabular}{l|ccccc}
        \Xhline{1.2pt}
        Window Size ($F_w \times H_w \times W_w$) ~~~~~~ & Param. (G) & Mem. (GB) & Time (ms) &  FVD $\downarrow$ & CLIPSIM $\uparrow$ \\
        \Xhline{0.4pt}
        $8\times1\times3$ & 1.601 & 10.07 & 149.42 & 525.91 & 0.3056 \\

        $4\times3\times6$ & 1.601 & 10.07 & 152.14 & 485.43 & 0.3064  \\
        $8\times3\times6$ (Final Setting) & 1.601 & 10.07 & 153.16 & 475.09 & 0.3107  \\
        $16\times3\times6$ & 1.601 & 10.07 & 153.23 & 487.08 & 0.3072  \\
        Global Attention & 1.601 & 23.51 & 205.58 & 485.86 & 0.3092  \\
        \Xhline{1.2pt}
    \end{tabular}
    \label{tab:aba_window_size}
\end{table*}

We further evaluate the effect of different window sizes.
The final window size is set to $8\times3\times6$, \ie, $F_w=8$, $H_w=3$, and $W_w=6$.
The rationale behind choosing $H_w = 3$ and $W_w = 6$ is to match the spatial resolution of the core feature in U-net, ensuring that the window attention in the core block can fully perceive the video contents.
As for $F_w$, we set it to 8 to achieve a broader temporal attention view while reducing computation complexity.
Table~\ref{tab:aba_window_size} shows the ablation study we performed on window sizes, following the experimental setup in Table~\ref{tab:aba_interaction} of our main paper.
Due to NVIDIA software differences, the memory values are not the same in Table~\ref{tab:aba_interaction} and Table~\ref{tab:aba_window_size}.
Our final configuration, $8\times3\times6$, achieves the best FVD, CLIPSIM scores and comparable efficiency.

\subsubsection{Video Generation Dataset}
\label{sec:abl_dataset}

\begin{figure}
    \centering
    \includegraphics[width=0.98\linewidth]{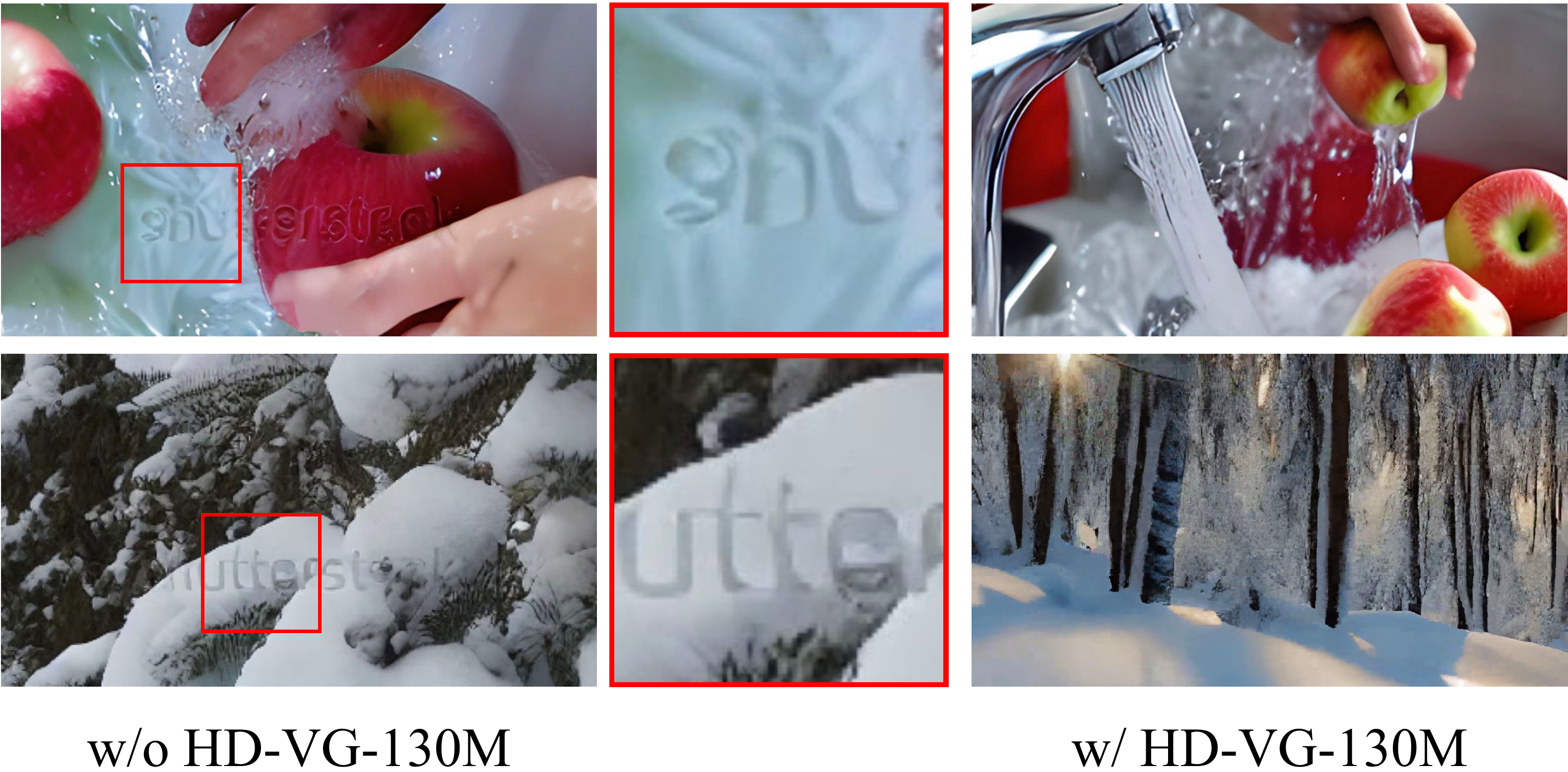}
    \vspace{2mm}
    \caption{Generation results without and with using HD-VG-130M for training the model.}
    \label{fig:com_w_wo_HD}
\end{figure}

\noindent\textbf{Visual Contents.}
{The advantages of HD-VG-130M extend beyond watermark removal.}
As shown in Table~\ref{tab:aba_dataset}, we evaluate the effect of our HD-VG-130M. After adding HD-VG-130M in training, the result on the validation set of WebVid-10M~\citep{bain2021frozen} has been improved by 45.34 in FVD, which verifies the superior quality of our HD-VG-130M for training text conditioned video generation model.
The visual comparison can also be found in Fig.~\ref{fig:com_w_wo_HD}. Training with HD-VG-130M not only eliminates watermarks but also elevates the scenic beauty and enriches the level of detail, leading to a comprehensive improvement in the visual quality of the generated videos.

\begin{table}[t]
    \centering
    \small \renewcommand{\arraystretch}{1.25}
    \caption{{Video generation effect of training on different datasets.}}
    \vspace{1mm}
    \label{tab:aba_dataset}
    \begin{tabular}{m{4.75cm} | c}
    \Xhline{1.2pt}
     Training Data        & FVD $\downarrow$ \\
    \Xhline{0.4pt}
     w/o  HD-VG-130M & 475.09 \\
     \Xhline{0.4pt}
     w/ HD-VG-130M & 429.75 \\
     \Xhline{0.4pt}
     {w/ HD-VG-130M + fine-tuning with higher-quality subset} & {418.40} \\
    \Xhline{1.2pt}
    \end{tabular}
\end{table}

\begin{figure}
    \centering
    \includegraphics[width=0.98\linewidth]{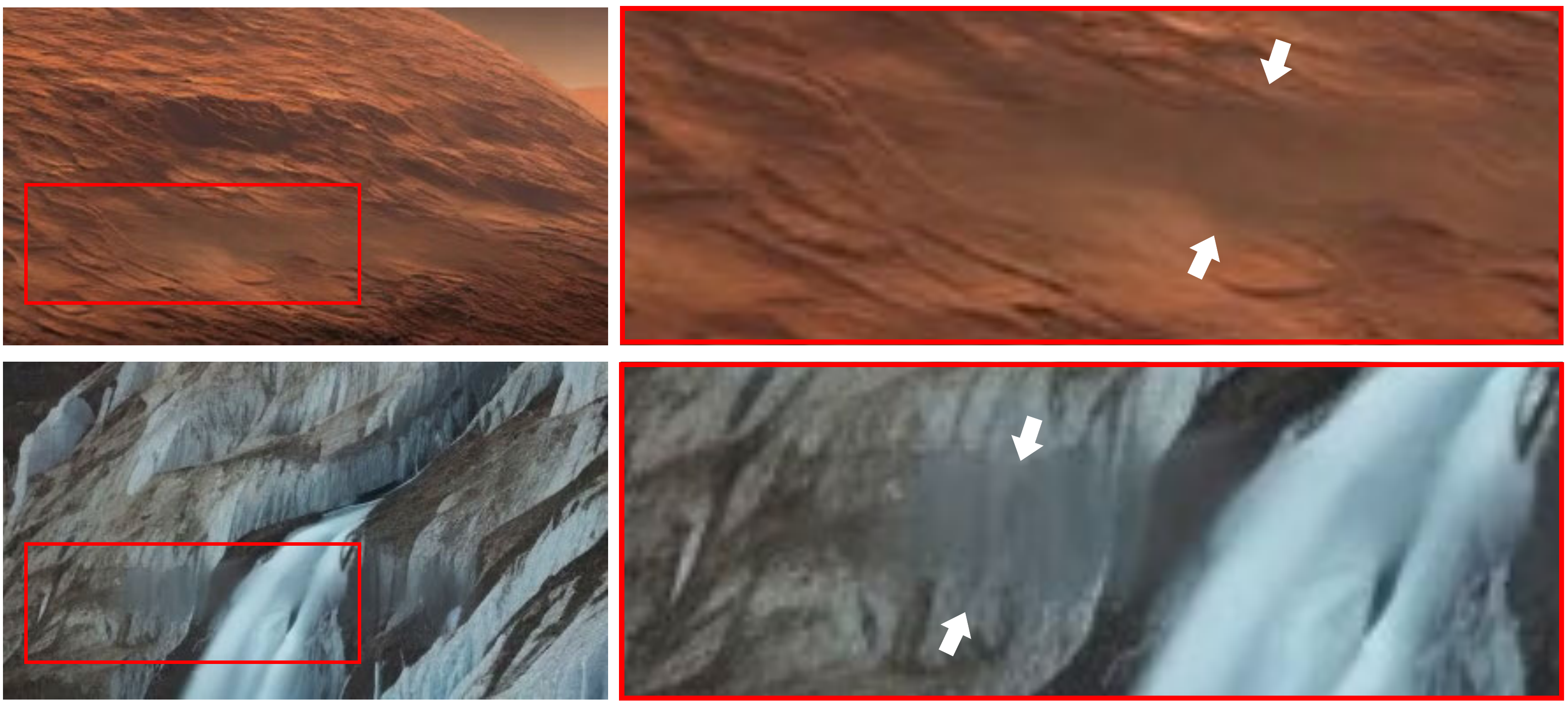}
    \vspace{2mm}
    \caption{{Results of the text-to-video model trained on de-watermarked WebVid-10M.}}
    \label{fig:com_video_dewatermarl}
\end{figure}

Regarding watermarks, we also tried using E2FGVI~\citep{E2FGVI} to remove watermarks from WebVid-10M.
As shown in Fig.~\ref{fig:com_video_dewatermarl}, the generated videos have blurry textures.
The locations of these blurry areas are in line with the locations of the original watermarks, indicating that the de-watermarking method causes blurriness, and this blurriness damages the training of the video generation model.
Removing watermarks from WebVid-10M to produce high-quality video data is non-trivial, which reveals the significance of our HD-VG-130M.

\begin{figure*}[t]
    \centering
    \includegraphics[width=0.99\linewidth]{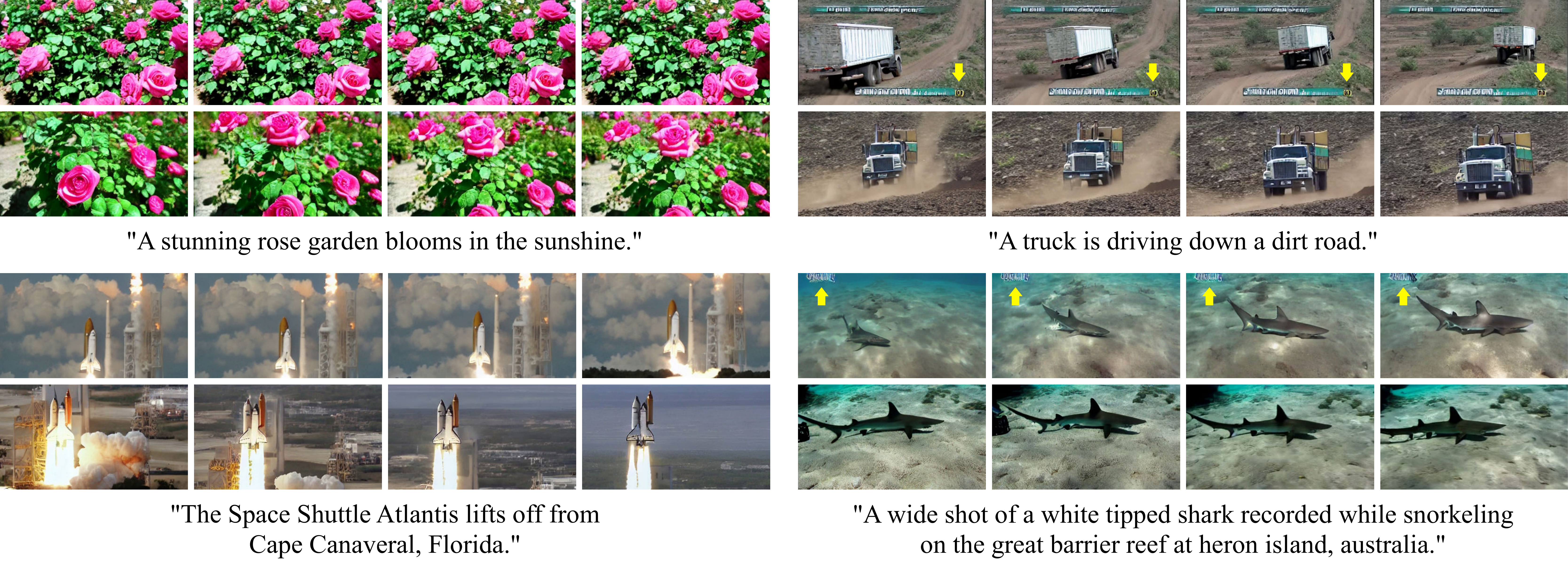}
    \vspace{3mm}
    \caption{Illustration comparing the impact of training with HD-VG-130M (top row in each group) and subsequent fine-tuning on the HD-VG-40M higher-quality subset (bottom row in each group). Yellow arrows indicate meaningless texts generated by training without HD-VG-40M.}
    \label{fig:aba_dataset}
\end{figure*}

\begin{figure}
    \centering
    \includegraphics[width=0.98\linewidth]{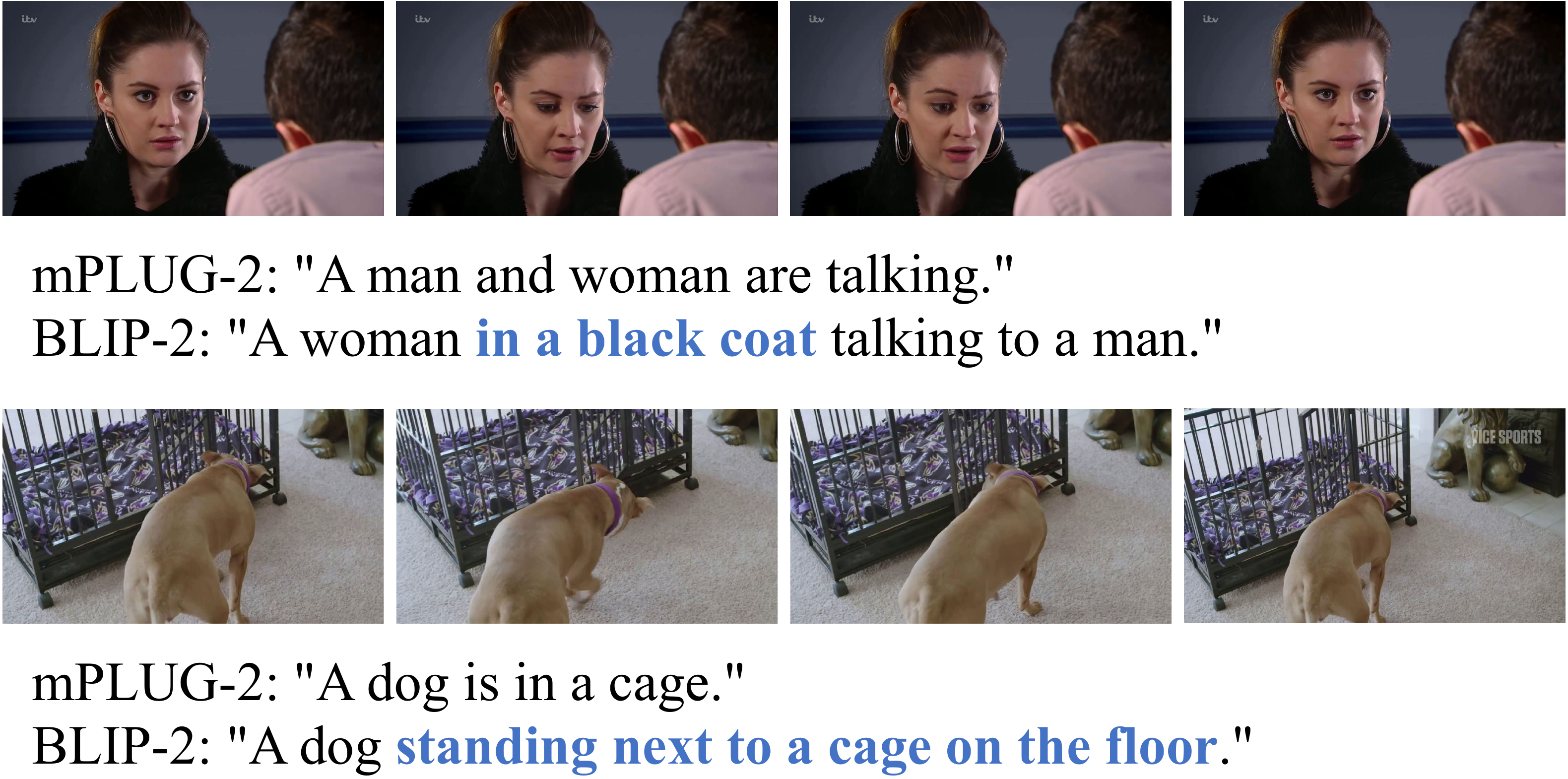}
    \vspace{2mm}
    \caption{{Comparison between using mPLUG-2 and BLIP-2 for annotating the contents of video.}}
    \label{fig:com_video_caption}
\end{figure}

Finally, we assess the effect of additional data processing. As shown in Table~\ref{tab:aba_dataset}, fine-tuning with the higher-quality subset enhances the FVD score to 418.40. Furthermore, visual comparisons are presented in Fig.~\ref{fig:aba_dataset}.
For the top-left sample, training without HD-VG-40M has a risk of generating static scenes, while for the bottom-left sample, the absence of HD-VG-40M in training leads to the space shuttle remaining stationary in each frame, essentially appearing as a translation transformation of a static image.
In the case of the right two samples, training without HD-VG-40M may generate meaningless text, as indicated by the yellow arrows. 
After fine-tuning on the higher-quality subset, these issues are resolved, and the aesthetics of the generated results improve, with better contrast, clearer edges, and more vivid colors.

\vspace{1mm}
\noindent\textbf{Video Captions.}
We further evaluated different captioning models.
We experimented with a state-of-the-art video captioning model, mPLUG-2~\citep{mPLUG2}, but observed that it provides less detailed descriptions (\eg, BLIP-2 predicts ``black coat" while mPLUG-2 does not in the first row of Fig.~\ref{fig:com_video_caption}) or misinterprets the scene (\eg, mistakes the dog to be inside the cage in the second row of Fig.~\ref{fig:com_video_caption}).
As a result, using videos captioned with mPLUG-2, the CLIPSIM is decreased to 0.3046.

In addition, we assessed the impact of training with HD-VILA-100M~\citep{xue2022advancing} instead of HD-VG-130M.
As HD-VILA-100M only provides subtitles and lacks scene detection (with potential multiple transitions), significant performance degradation is observed in FVD (429.75 $\rightarrow$ 692.99) and CLIPSIM (0.3082 $\rightarrow$ 0.2671), despite joint training with WebVid. This experiment highlights the crucial role of our scene detection and video captioning procedures.

\begin{table}[t]
    \centering
    \small \renewcommand{\arraystretch}{1.25}
    \caption{Comparison of text-to-video generation performance on the UCF101 dataset.}
    \vspace{1mm}
    \label{tab:result_ucf101}
    \begin{tabular}{l|cc}
        \Xhline{1.2pt}
        Method       & Zero-shot & FVD$\downarrow$    \\ 
        \Xhline{0.4pt}
        VideoGPT~\citeyearpar{yan2021videogpt}     & No        & 2880.6 \\
        MoCoGAN~\citeyearpar{2018MoCoGAN}      & No        & 2886.8 \\
        MoCoGAN-SG2~\citeyearpar{skorokhodov2022stylegan}   & No        & 1821.4 \\
        MoCoGAN-HD~\citeyearpar{tian2021good_mocohd}   & No        & 1729.6 \\
        DIGAN~\citeyearpar{yu2022generating_digan}        & No        & 1630.2 \\
        StyleGAN-V~\citeyearpar{skorokhodov2022stylegan}   & No        & 1431.0 \\
        PVDM~\citeyearpar{PVDM}         & No        & 343.6  \\ 
        \Xhline{0.4pt}
        CogVideo~\citeyearpar{2023CogVideo}     & Yes       & 701.6  \\
        MagicVideo~\citeyearpar{zhou2022magicvideo}   & Yes       & 699.0  \\
        LVDM~\citeyearpar{he2022lvdm}   & Yes       & 641.8  \\
        ModelScope~\citeyearpar{VideoFusion}   & Yes       & 639.9  \\
        Video LDM~\citeyearpar{VideoLDM}    & Yes       & 550.6  \\ 
        LaVie~\citeyearpar{LAVIE}    & Yes       & 526.3 \\
        AnimateDiff~\citeyearpar{guo2023animatediff} & Yes & 499.3 \\
        AnimateDiff+Panda~\citeyearpar{panda70M} & Yes & 421.9 \\
        \Xhline{0.4pt}
        Ours w/o FT         & Yes       & 410.0      \\ 
        Ours w/ FT          & Yes       & {\textbf{398.1}}       \\ 
        \Xhline{1.2pt}
    \end{tabular}
\end{table}

\subsection{Quantitative and Qualitative Comparison}
\label{sec:comparison_results}

To thoroughly evaluate the performance of our VideoFactory, we benchmark it on three distinct datasets: the WebVid-10M~\citep{bain2021frozen} (Val) dataset, which shares the same domain as part of our training data, as well as the UCF101~\citep{soomro2012ucf101} and the MSR-VTT~\citep{xu2016msr} {datasets} in a zero-shot setting.
We also demonstrated results with and without fine-tuning on the HD-VG-40M higher-quality subset, denoted as ``Ours w/o FT'' and ``Ours w/ FT'' respectively.

\vspace{1mm}
\noindent \textbf{Evaluation on UCF101.} 
As mentioned in Sec.~\ref{sec:hd_vg}, the textual annotations in UCF101 are class labels. We first follow~\citep{2022_VDM, MakeAVideo} and rewrite the labels of 101 classes to descriptive captions, and then generate 100 samples for each class.
As shown in Table~\ref{tab:result_ucf101}, the FVD of our methods reaches {398.1}, which achieves the best compared with other methods both in zero-shot setting and beats most of the methods which have tuned on UCF101.
The results verify that our proposed VideoFactory could generate more coherent and realistic videos.

\begin{figure*}[t]
    \centering
    \includegraphics[width=0.98\linewidth]{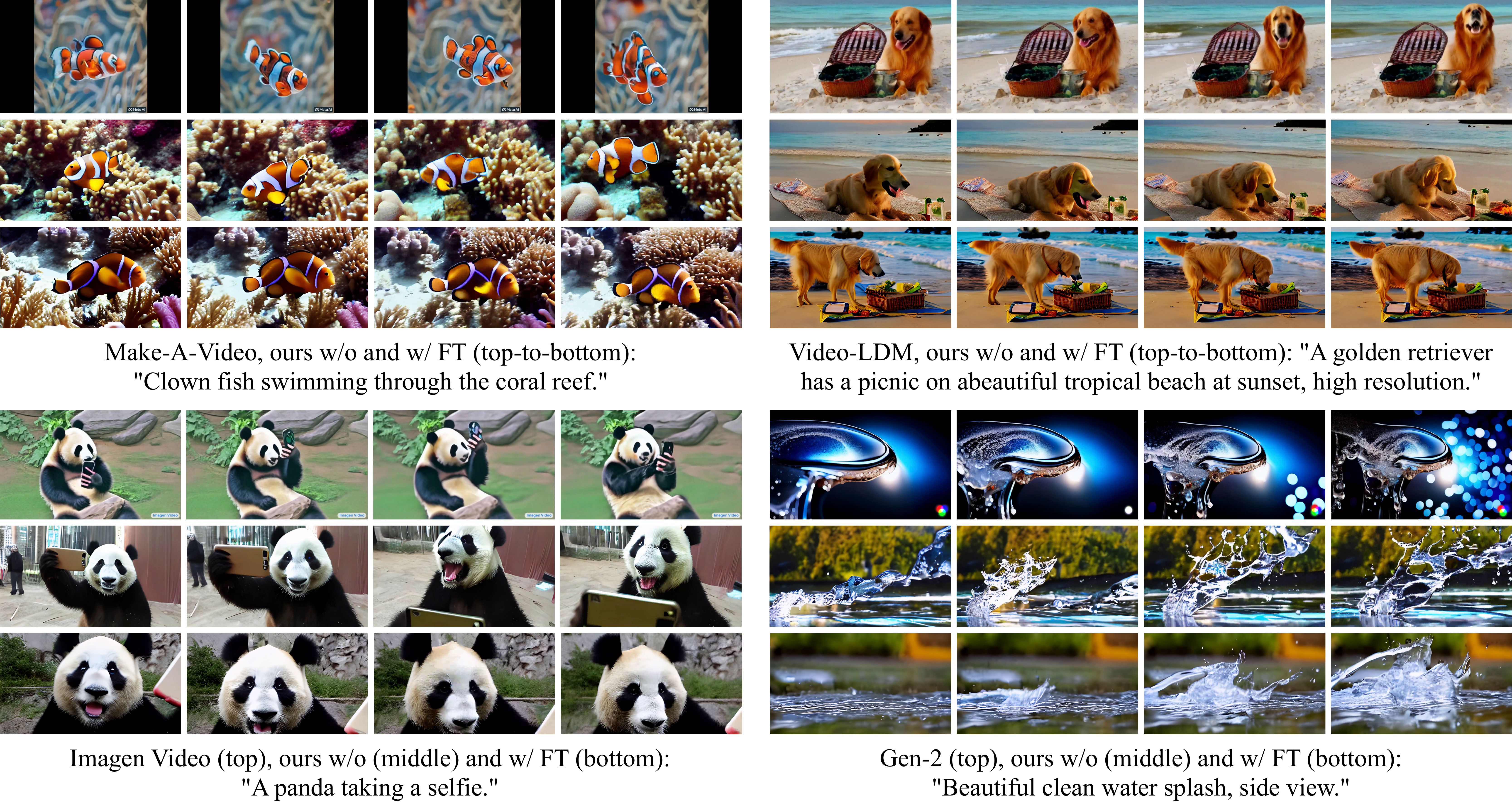}
    \vspace{2mm}
    \caption{Text-to-video generation results compared with Make-A-Video, Imagen Video, Video-LDM, and Gen-2 (Cases of the first three methods are collected from their public project websites).}
    \label{fig:com_imagen_makeavideo}
\end{figure*}

\begin{table}[t]
    \centering
    \small \renewcommand{\arraystretch}{1.25}
    \caption{Comparison of text-to-video generation performance on the MSR-VTT dataset.}
    \vspace{1mm}
    \label{tab:result_msrvtt}
    \begin{tabular}{m{3.3cm}|cc}
        \Xhline{1.2pt}
        Method       & Zero-shot & CLIPSIM$\uparrow$ \\ 
        \Xhline{0.4pt}
        GODIVA~\citeyearpar{wu2021godiva}       & No        & 0.2402  \\
        NUWA~\citeyearpar{wu2022nuwa}         & No        & 0.2439  \\
        \Xhline{0.4pt}
        LVDM~\citeyearpar{he2022lvdm}   & Yes       & 0.2381 \\
        CogVideo~\citeyearpar{2023CogVideo}     & Yes       & 0.2631  \\
        ModelScope~\citeyearpar{VideoFusion}    & Yes       & 0.2795  \\
        AnimateDiff~\citeyearpar{guo2023animatediff} & Yes & 0.2869 \\
        AnimateDiff +Panda~\citeyearpar{panda70M} & Yes & 0.2880 \\
        Video LDM~\citeyearpar{VideoLDM}    & Yes       & 0.2929  \\ 
        LaVie~\citeyearpar{LAVIE}    & Yes       & 0.2949 \\
        \Xhline{0.4pt}
        Ours w/o FT          & Yes       & 0.3005       \\ 
        Ours w/ FT          & Yes       & \textbf{0.3021}       \\ 
        \Xhline{1.2pt}
    \end{tabular}
\end{table}

\vspace{1mm}
\noindent \textbf{Evaluation on MSR-VTT.} As shown in Table~\ref{tab:result_msrvtt}, we also evaluate the CLIPSIM on the widely used video generation benchmark MSR-VTT. We randomly choose one prompt per example from MSR-VTT to generate 2990 videos in total. Although in a zero-shot setting, our method achieves the best compared to other methods with an average CLIPSIM score of 0.3021, which suggests the semantic alignment between the generated videos and the input text.
Moreover, note that the state-of-the-art AnimateDiff~\citeyearpar{guo2023animatediff} training on Panda~\citeyearpar{panda70M} performs inferior to ours for both FVD on UCF101 and CLIPSIM on MSR-VTT, demonstrating the effectiveness of both our dataset and model designs.

\begin{table}[t]
    \small \renewcommand{\arraystretch}{1.25}
    \caption{Comparison of text-to-video generation performance on the WebVid dataset.}
    \vspace{1mm}
    \label{tab:results_WebVid-10M}
    \centering
    \begin{tabular}{l|cc}
        \Xhline{1.2pt}
        Method           & FVD$\downarrow$ & CLIPSIM$\uparrow$ \\ 
        \Xhline{0.4pt}
        LVDM~\citeyearpar{he2022lvdm}     & 455.53 &  0.2751 \\ 
        ModelScope~\citeyearpar{VideoFusion}        & 414.11 &  0.3000 \\
        \Xhline{0.4pt}
        Ours w/ FT            & \textbf{322.13} & \textbf{0.3104} \\ 
        \Xhline{1.2pt}
    \end{tabular}
\end{table}

\vspace{1mm}
\noindent \textbf{Evaluation on WebVid-10M (Val).} 
Referring to Table~\ref{tab:results_WebVid-10M}, we randomly extract 5K text-video pairs from WebVid-10M which are exclusive from the training data to form a validation set and conduct evaluations on it.
Our approach achieves an FVD of 292.35 and a CLIPSIM of 0.3070, outperforming existing methods and showcasing the superiority of our approach.

\begin{figure*}[t]
    \centering
    \includegraphics[width=0.98\linewidth]{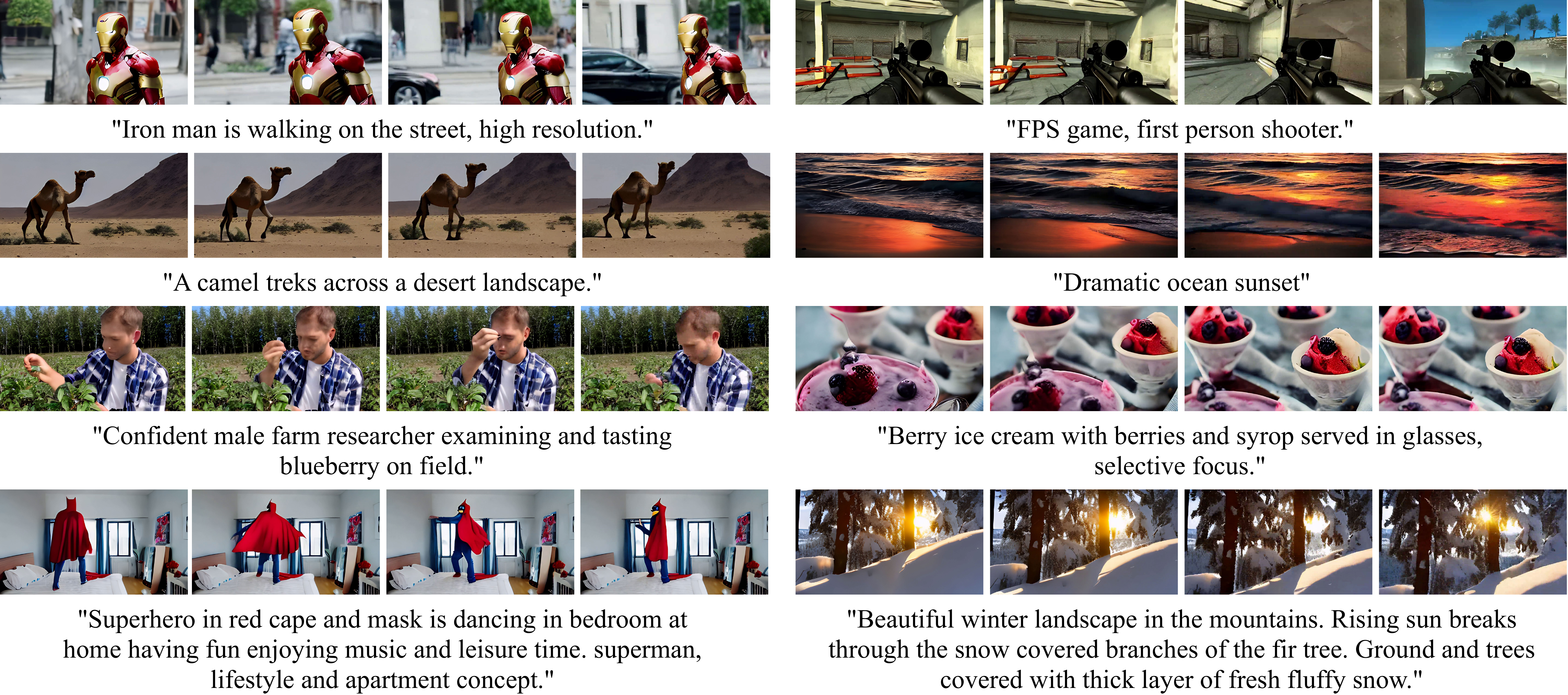}
    \vspace{2mm}
    \caption{Samples generated by our VideoFactory (w/ FT) exhibit high quality, featuring clear motion, intricate details, and precise semantic alignment.}
    \label{fig:showcase_our}
\end{figure*}

\vspace{1mm}
\noindent \textbf{Subjective Results.}
In Fig.~\ref{fig:com_imagen_makeavideo}, we show comparison results against Make-A-Video, Imagen Video, and Video LDM. The prompts and generated results are collected from their official project website.
We also evaluate Gen-2~\footnote{\url{https://research.runwayml.com/gen2}}, a popular platform in the AIGC field.
Make-A-Video only generates 1:1 videos, which limits the user experience. 
When compared with Imagen Video and Video LDM, our model generates the panda and golden retriever with more vivid details.
Despite setting the motion intensity parameter to the maximum, Gen-2 cannot simulate the splashing motion of water. We showcase additional samples of our model in Fig.~\ref{fig:showcase_our} and more in the \textit{supplementary}.

\begin{figure}[t]
    \centering
    \includegraphics[width=0.98\linewidth]{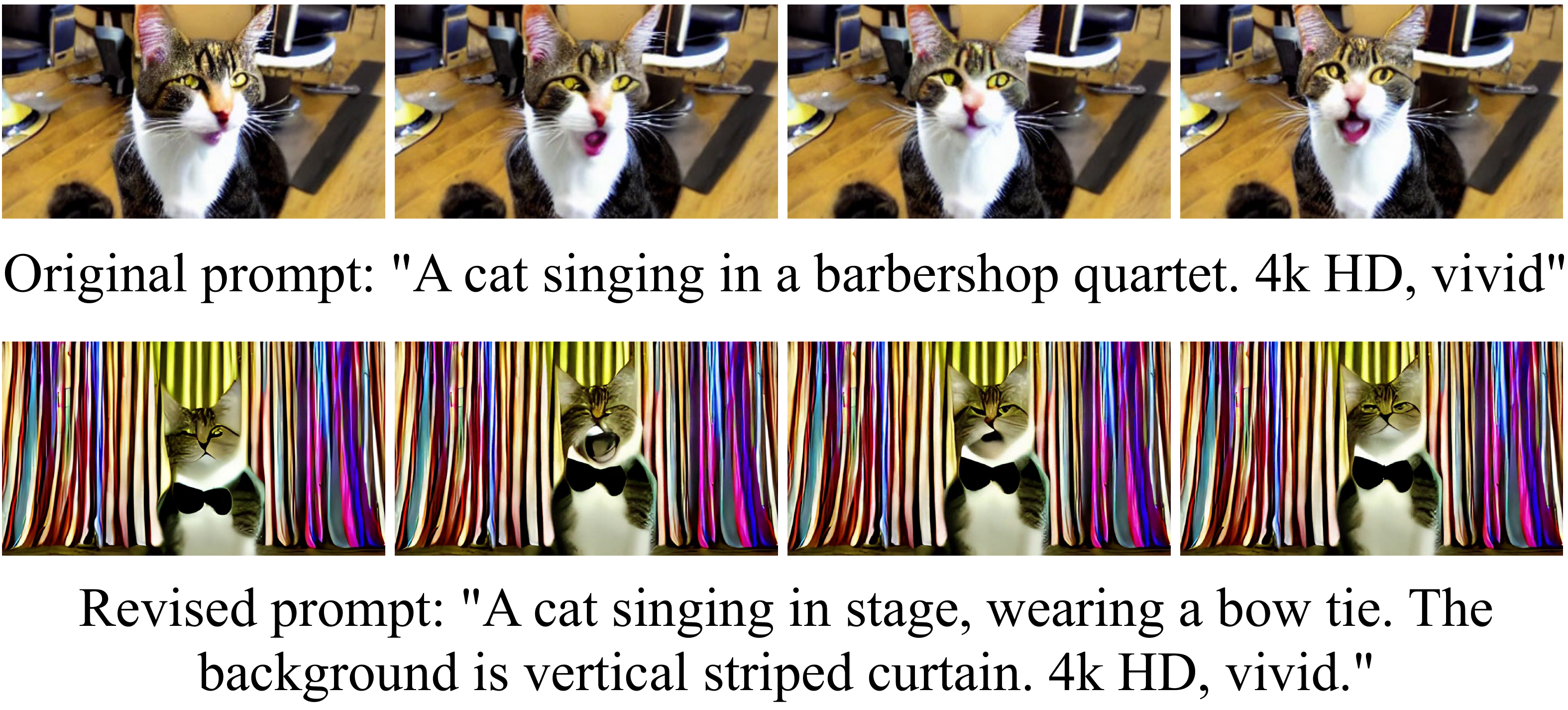}
    \vspace{3mm}
    \caption{Failure case study of our text-to-video generation model and a quick solution.}
    \label{fig:failure_model}
\end{figure}

\vspace{1mm}
\noindent \textbf{Failure Case Study.}
The typical failure case of our text-to-video generation model is that our text encoder, CLIP~\citep{CLIP}, can sometimes misinterpret concepts, leading to unintended results. For instance, with the input prompt ``A cat singing in a barbershop quartet," the term ``barbershop quartet" signifies musical performance in a specific style.
However, our text encoder might inadvertently emphasize ``barbershop", introducing a corresponding background to the video. To address this, we can use GPT-3.5 for prompt refinement, after which our model can generate a vivid cat singing on the stage. A visual demonstration (we use the w/o FT version for convenience) can be found in Fig.~\ref{fig:failure_model}.

\section{Conclusion}
\label{sec:conclusion}
In this paper, we introduce a high-quality open-domain video generation framework that produces watermark-free, high-definition, widescreen videos.
We enhance spatial and temporal modeling using a novel swapped cross-attention mechanism, allowing spatial and temporal information to complement each other effectively.
Additionally, we provide the HD-VG-130M dataset, featuring 130 million open-domain text-video pairs in widescreen, watermark-free, high-definition format, maximizing the potential of our model. 
A higher-quality subset is constructed to further promote the performance.
Experimental results demonstrate that our method generates videos with superior spatial quality, temporal consistency, and alignment with text.
Analysis also demonstrates the effectiveness of our dataset and processing designs. 

Future directions for our work may involve refining BLIP-2 captions using large language models and changing the backbone to more powerful text-to-image generation baselines. The field of video generation has experienced significant growth recently. Due to limited resources, we cannot match the capabilities of some closed-source industrial products. However, we believe that our contributions, particularly the open-source dataset and comprehensive experimental analysis, will benefit the advancement of this field.

\bibliography{sn-bibliography}

\end{document}